\documentclass[conference]{IEEEtran}
\IEEEoverridecommandlockouts
\usepackage{cite}
\usepackage{amsmath,amssymb,amsfonts}
\usepackage{algorithmic}
\usepackage{graphicx}
\usepackage{textcomp}
\usepackage{xcolor}
\usepackage{booktabs,nicematrix,makecell}
\usepackage{xparse}
\usepackage{multirow}
\usepackage{lscape}
\usepackage{graphicx}
\usepackage{pifont}
\usepackage{tcolorbox}
\usepackage{booktabs}
\usepackage{hyperref}
\usepackage{enumitem}
\usepackage{tabularx}
\usepackage{colortbl}
\usepackage{array}
\usepackage{wasysym}
\usepackage{cleveref}
\usepackage{acronym}

\newcommand{\para}[1]{\vskip 4pt\noindent\textbf{#1}\hskip .05in}

\newenvironment{highlight}[1][]
{
    \begin{tcolorbox}[
        colback=gray!15, 
        colframe=gray!80, 
        rounded corners, 
        arc=3mm, 
        top=1.5mm, 
        bottom=1.5mm, 
        left=3mm, 
        right=3mm, 
        #1 
    ]
}
{
    \end{tcolorbox}
}

\def\BibTeX{{\rm B\kern-.05em{\sc i\kern-.025em b}\kern-.08em
    T\kern-.1667em\lower.7ex\hbox{E}\kern-.125emX}}
\begin{document}

\title{SoK: What Makes Private Learning Unfair?
\thanks{This work has been accepted for publication in the 3rd IEEE Conference on Secure and Trustworthy Machine Learning (SaTML'25). The final version will be available on IEEE Xplore.}
}


\author{
\IEEEauthorblockN{Kai Yao}
\IEEEauthorblockA{\textit{School of Informatics}\\
\textit{University of Edinburgh}\\
\href{mailto:kai.yao@ed.ac.uk}{kai.yao@ed.ac.uk} \\}
\and
\IEEEauthorblockN{Marc Juarez}
\IEEEauthorblockA{\textit{School of Informatics}\\
\textit{University of Edinburgh}\\
\href{mailto:marc.juarez@ed.ac.uk}{marc.juarez@ed.ac.uk} \\}
}

\maketitle

\begin{abstract}
Differential privacy has emerged as the most studied framework for privacy-preserving machine learning. However, recent studies show that enforcing differential privacy guarantees can not only significantly degrade the utility of the model, but also amplify existing disparities in its predictive performance across demographic groups. Although there is extensive research on the identification of factors that contribute to this phenomenon, we still lack a complete understanding of the mechanisms through which differential privacy exacerbates disparities. The literature on this problem is muddled by varying definitions of fairness, differential privacy mechanisms, and inconsistent experimental settings, often leading to seemingly contradictory results.

This survey provides the first comprehensive overview of the factors that contribute to the disparate effect of training models with differential privacy guarantees. We discuss their impact and analyze their causal role in such a disparate effect. Our analysis is guided by a taxonomy that categorizes these factors by their position within the machine learning pipeline, allowing us to draw conclusions about their interaction and the feasibility of potential mitigation strategies. We find that factors related to the training dataset and the underlying distribution play a decisive role in the occurrence of disparate impact, highlighting the need for research on these factors to address the issue.
\end{abstract}

\begin{IEEEkeywords}
machine learning, differential privacy, privacy-preserving ML, fairness, trustworthy ML
\end{IEEEkeywords}

\section{Introduction}

The widespread adoption of Machine Learning (ML) has raised serious privacy concerns, especially in sectors that handle sensitive personal data, such as healthcare, finance, and criminal justice. To address these concerns, researchers have proposed privacy-preserving techniques that can seamlessly integrate with existing ML pipelines.
The most popular privacy paradigm among these researchers is Differential Privacy (DP) \cite{dwork2006differential}, which has served as the foundation for several privacy-preserving ML (PPML) algorithms, including the popular Differentially Private Stochastic Gradient Descent (DP-SGD) \cite{abadi2016deep}.
However, although DP offers formal privacy guarantees for the individuals whose data are used to train the models, it often comes at the cost of reduced model utility.

In parallel, recent studies have shown that the repercussions of training ML models with DP guarantees go beyond merely reducing the utility of the model. Bagdasaryan et al.\ were the first to observe that models trained with DP-SGD not only suffer a reduction in utility, but that the reduction is unevenly distributed across demographic groups, with underrepresented groups suffering a disproportionate decrease in utility compared to other groups \cite{bagdasaryan2019differential}. The authors confirm this phenomenon by applying DP-SGD in various learning tasks, including sentiment analysis and face detection, suggesting that its cause lies in the internal workings of DP-SGD.

These observed effects of DP training can have harmful impacts on minorities and underrepresented groups. These groups have already been shown to suffer from the disparate performance of ML models in a wide range of learning tasks and applications \cite{buolamwini2018gender,sap2019risk,celi2022sources,mehrabi2021survey}.
However, if DP becomes the de-facto standard for PPML, as current trends suggest, its potential to exacerbate existing disparities could have a tremendous impact across many domains. This impact is particularly concerning in areas like employment, education, and lending, where decision-making bias can undermine the life prospects of affected individuals and perpetuate societal inequities~\cite{barocas2016big}.

Among the early efforts to understand how DP exacerbates the disparities, researchers have identified the gradient clipping mechanism of DP-SGD as one of the main contributing factors~\cite{bagdasaryan2019differential,xu2020removing,tran2021differentially,esipova2022disparate,srivastava2024amplifying}. Consequently, mitigation strategies have focused on modifying the clipping mechanism to minimize its disparate impact~\cite{xu2020removing,tran2021differentially,esipova2022disparate,liu2022mitigating}. Research also extends beyond DP-SGD to explore alternative DP algorithms within the DP framework~\cite{uniyal2021dp,xu2019achieving,carey2023randomized,liu2022mitigating,mangold2023differential,tran2021differentially,makhlouf2024impact,noe2022exploring,makhlouf2024systematic}, some of which exhibit less disparate impact than DP-SGD. Furthermore, several studies have identified factors external to the DP technique itself that may contribute to its exacerbation effects. These include the model capacity and architecture \cite{mangold2023differential,de2023empirical}, training hyperparameters \cite{de2023empirical}, and the characteristics of the training dataset \cite{mangold2023differential,cummings2019compatibility,arasteh2023private,tran2021differentially}. The role of group imbalance \cite{xu2020removing,sanyal2022unfair,farrand2020neither} and the difference in distance to the model's decision boundary between groups~\cite{mangold2023differential,tran2021differentially,xu2019achieving} have also been studied in the literature.
Despite these preliminary efforts, this problem remains largely unexplored, leaving a gap in our understanding of how differential privacy causes these disparities.

This survey provides the first comprehensive overview of all the factors that have been identified to contribute to DP-induced disparities in PPML, categorizing them in a novel taxonomy that helps researchers reason about the practical implications of the occurrence of a factor. Furthermore, we analyze the causal roles these factors play in the incidence of a disparity, allowing us to draw conclusions about their connection to the observed phenomenon and propose potential research directions for addressing the issue.

Fioretto et al.\ have previously surveyed the interactions between privacy and fairness in decision-making algorithms in general, including PPML~\cite{fioretto2022differential}. In contrast, our survey exclusively focuses on the literature on PPML, a rapidly growing field that has recently uncovered several important contributing factors that are not included in Fioretto et al.'s study. More importantly, our work not only extends prior work by incorporating recent research on the problem but also systematizes it around a novel taxonomy, allowing us to discover relationships among various factors and critically assess the evidence supporting each factor's contribution and causal necessity to the exacerbation of disparities.

\medskip
\noindent Our main contributions include:

\para{A comprehensive overview of contributing factors.} We systematically compile and discuss the factors identified in the literature that contribute to DP-induced disparity exacerbation.

\para{A novel taxonomy of factors.} We propose a taxonomy that categorizes factors based on their position within the ML pipeline. This taxonomy also serves as a foundation for informing the design of effective mitigation strategies.

\para{A causal analysis.} We conduct a causal analysis grounded in existing evidence to identify potential cause-and-effect relationships between the factors and DP-induced disparity exacerbation. Where applicable, we provide reasoning for the necessity and sufficiency of specific factors in contributing to this exacerbation. Our findings indicate that the combination of a small training dataset and significant disparities in the distance to the decision boundary across groups is the only condition that may be sufficient to trigger the exacerbation. 

\para{A discussion of countermeasures.} We overview the existing mitigation strategies based on the taxonomy, highlighting their limitations to help researchers and practitioners develop more practical and effective mitigation solutions.

\para{An overview of open challenges.} We identify research gaps in this field and offer insights into future research directions.


\section{Preliminaries}
\label{sec:preliminaries}
This section offers the background needed to understand the remainder of the survey, including an introduction to supervised learning and definitions of privacy and fairness within the context of supervised learning.

\subsection{Supervised Learning}
Our study centers on \emph{supervised learning} for classification, as it is the primary focus of research at the intersection of PPML and ML fairness.

In supervised learning, the aim is to learn a function $f$ that maps an input $x\in X$ to an output $y\in Y$ given an underlying probability distribution over $X$ and $Y$ \cite{shalev2014understanding}. This is achieved through a labeled dataset $\mathcal{D} = \{(x_i, y_i)\}_{i=1}^N$, where $N$ is the dataset size, and each $(x_i, y_i)$ is independently drawn from the joint distribution of $X$ and $Y$. The goal is to use $\mathcal{D}$ to find an $h_\theta$, called the \emph{model}, which approximates the true mapping $f$ from $X$ to $Y$.

Within the Empirical Risk Minimization (ERM) paradigm, we find the optimal parameters $\theta^*$ by minimizing a predefined loss function $L$ on $\mathcal{D}$, guided by an optimization algorithm such as Stochastic Gradient Descent (SGD)~\cite{bottou1991stochastic}. We refer to the process of finding the optimal parameters as \emph{training} the model and, consequently, we call $\mathcal{D}$ the \emph{training dataset}.

More formally, ERM is formulated as solving the following optimization problem:
\begin{equation}\label{eq:erm}
\theta^* := \min_{\theta} \mathbb{E}_{(x, y) \sim P(X, Y)} \left[ L(y, h_\theta(x)) \right],
\end{equation}

\noindent
where $L$ penalizes the prediction errors of $h_{\theta}(x)$.

The expected loss in \Cref{eq:erm} is estimated on $\mathcal{D}$ to obtain the empirical loss. When $\mathcal{D}$ is small, the resulting model is prone to overfitting the training data. To improve generalization, a common practice is to \emph{regularize} the learning problem by adding constraints to it. For example, a common regularization technique is to add penalty terms to the objective function that discourages complex models.

In practice, the predictive performance of $h_\theta$ is also influenced by hyperparameters that adjust aspects of the training process, including parameters of the optimization algorithm.

\subsection{Privacy in Supervised Learning}
\label{privacy_in_ml}
Privacy in supervised learning often refers to protecting sensitive information in the training dataset, including information about the individuals who contributed to the training dataset with their personal data.
There are various existing data privacy frameworks, but most PPML algorithms fall within DP, a rigorous mathematical framework that controls the risk of revealing whether an individual's data is included in the training dataset \cite{dwork2006differential,kasiviswanathan2011can}. 
Due to the popularity of DP as the privacy notion to aim for in supervised learning, in this work we focus on DP-based PPML techniques.

\para{Differential Privacy (DP).} 
Formally, a randomized algorithm $\mathcal{A}$ (e.g., an algorithm to solve the optimization problem in \Cref{eq:erm} satisfies $(\varepsilon, \delta)$-DP if for any pair of neighboring datasets $\mathcal D$ and $\mathcal D'$ differing in at most one data point, and for any measurable subset $S$ of the range of $\mathcal A$,
\begin{equation}\label{eq:dp}
\text{Pr}[\mathcal{A}(\mathcal D) \in S] \leq e^\varepsilon \cdot \text{Pr}[\mathcal{A}(\mathcal D') \in S] + \delta,
\end{equation}

\noindent
where $\varepsilon > 0$ is the privacy parameter controlling the privacy loss and $\delta > 0$ quantifies the probability of failure in satisfying $\varepsilon$-DP. This inequality implies that the probability of any outcome $S$ under algorithm $\mathcal{A}$ remains approximately the same under the removal or addition of any single point to the data, with probability $\delta$ of the inequality not holding. Essentially, this condition limits the information that an adversary could obtain from observing $\mathcal A$'s output about the membership of any individual data point to the training dataset.

In the context of supervised learning, DP can be incorporated into various stages of the ML pipeline, such as data collection, model training, and publishing of model outputs \cite{ponomareva2023dp}. Local DP (LDP), which protects individual data points before applying $\mathcal A$, is typically implemented during data collection, while global DP (GDP), which aggregates and releases the protected result, is implemented at later stages of the ML pipeline. If DP is satisfied in the supervised learning setting, the effect of the presence or absence of any single data point on the model's output is bounded. As a result, the model enables users to extract valuable insights from datasets containing personal data while minimizing the risk of revealing whether an individual's data is included in the training set.

\para{Differentially Private SGD (DP-SGD).} DP-SGD currently stands as the most studied privacy algorithm to ensure DP guarantees on models trained via ERM~\cite{abadi2016deep}. DP-SGD integrates the original SGD optimization technique with the principles of DP, including a method to keep track of the privacy loss in each step of SGD, thus providing tight DP bounds~\cite{abadi2016deep}.
DP-SGD achieves $(\varepsilon, \delta)$-DP through the combination of two mechanisms: noise addition and gradient clipping. Noise addition follows standard DP mechanisms by iteratively injecting the noise into the gradients in each step of SGD. In order to bound the sensitivity of the training procedure, and thus limit the amount of added noise, gradient clipping ensures a bound on the norm of the gradients calculated for each SGD step.

\subsection{Fairness in Supervised Learning}
\label{fairness_notions}
In the context of supervised ML models, the literature provides statistical definitions of \emph{fairness} in relation to model outcomes~\cite{barocas2023fairness}. These definitions are generally categorized into two broad notions: \emph{individual fairness} and \emph{group fairness}. Individual fairness imposes that similar individuals receive similar treatment by the model. Group fairness requires that model performance should be equitable across demographic groups defined by protected attributes such as race, gender, or socioeconomic status.
DP aligns, by definition, with individual fairness~\cite{dwork2012fairness}, while its exacerbation of disparities is observed for group fairness; hence, our survey focuses on the latter.

\para{Group fairness.} Given a model \( h_\theta \) and two groups \( A \) and \( B \), the unfairness against \( B \) can be quantified by:
\begin{equation}
\mathcal{M}(h_\theta; A) - \mathcal{M}(h_\theta; B)
\end{equation}
where \( \mathcal{M}(h_\theta; X) \) represents a \emph{performance metric} \( \mathcal{M} \) applied to the model \( h_\theta \) and group \( X \). This formula captures key group fairness notions in the literature, including:
\begin{enumerate}[left=0pt]
    \item \textbf{Accuracy Parity} (AP): \( \mathcal{M} := \text{Accuracy} \)  \cite{zafar2017fairness},
    \item \textbf{Demographic Parity} (DEP): \( \mathcal{M} := \text{Positivity Rate} \) \cite{mehrabi2021survey},
    \item \textbf{Equal Opportunity} (EOP): \( \mathcal{M} := \text{TPR} \) \cite{hardt2016equality},
    \item \textbf{Equalized Odds} (EOD): \( \mathcal{M} := \text{TPR} \land \text{FPR} \) \cite{hardt2016equality},
    \item \textbf{Predictive Equality Parity} (PEP): \( \mathcal{M} := \text{FPR} \) \cite{corbett2017algorithmic},
    \item \textbf{Predictive Rate Parity} (PRP): \( \mathcal{M} := \text{Precision}\) \cite{chouldechova2017fair},
    \item \textbf{Risk Parity} (RP): \( \mathcal{M} := \text{Empirical Risk}\) \cite{tran2021differentially}.
\end{enumerate}

Many of the research articles included in our survey focus exclusively on accuracy parity (1), while some others explore multiple fairness notions simultaneously, typically considering a binary classification setting and metrics specific to this setting (2-6). Another line of research focuses on analyzing empirical risk parity (7).

\section{Scope and Methodology}
\label{sec:scope}

Our investigation focuses on scientific articles that satisfy \emph{both} of the following conditions:

\begin{enumerate}
    \item They apply \textbf{DP} to \textbf{supervised ML} processes, whether during training or other stages of the ML pipeline.
    \item They evaluate \textbf{fairness} issues by examining the model \textbf{performance disparity} across different groups within the data, before and after DP, and/or by studying strategies that mitigate such disparity.
\end{enumerate}

We restrict our survey to papers that contribute to identifying and evaluating contributing factors, either empirically or theoretically, because our aim is to systematize the body of work and advance our understanding of the role these factors play in the unfairness exacerbation effect of DP.
We exclude tangentially related studies that do not directly address the issue.
For example, we are aware of several works that explore the simultaneous achievement of privacy and fairness by integrating DP into fair learning techniques~\cite{ghoukasian2024differentially, jagielski2019differentially}. Although valuable for the broader field, these works do not study the impact of DP on \emph{exacerbating unfairness}; therefore, we consider them out of the scope of this survey.

We submitted the search query in Table~\ref{tab:googlescholarquery} (Appendix~\ref{appendix_googlescholar}) to \href{https://scholar.google.com/}{Google Scholar} to populate the initial pool of articles. This query returned over 200 papers, each of which was reviewed to assess whether it fell within the scope of our survey. 
We selected 20 studies from this pool for further review. The selected studies were systematically examined for key elements such as fairness notions, DP techniques, ML algorithms, tasks and data modalities, and factors identified as contributing to the unfairness issue brought about by DP.



\section{Factors Contributing to Unfairness}

\label{sec:factors}
\begin{figure}[htbp]
    \centering
    \includegraphics[width=0.45\textwidth]{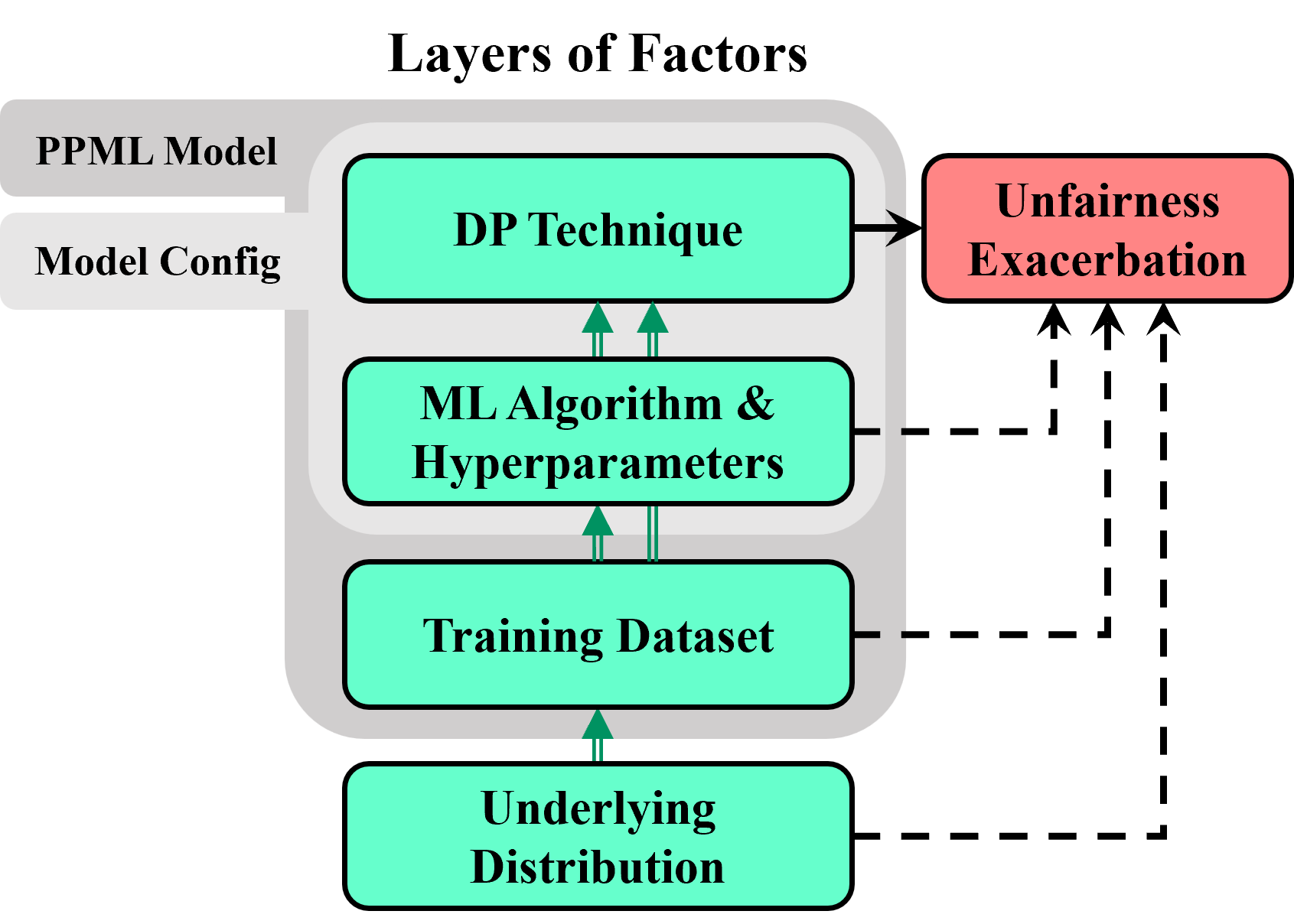}
    \caption{Taxonomy of factors contributing to DP's exacerbation of unfairness in PPML. The factors are divided into four layers (green boxes) according to the stage of the ML pipeline they belong to. The top three layers are grouped together to convey that they can be adjusted at training time. The top two layers are further grouped as components of a PPML model's configuration. The double green arrows convey cross-layer relationships: the DP technique is selected based on the training dataset and the ML algorithm \& hyperparameters, while the latter are chosen based on the training dataset; and the training dataset is sampled from the underlying distribution. All four layers contain factors that contribute to unfairness exacerbation, with the DP layer being the direct cause (solid black arrow) and the other layers having a rather indirect impact (dashed black arrows).}
    \label{fig:taxonomy}
\end{figure}

Recent research suggests that performance disparities arising from the application of PPML techniques cannot be solely attributed to the DP techniques themselves, and that there are other factors involved.
However, there is no comprehensive taxonomy for categorizing and analyzing these factors. We propose organizing these contributing factors into four distinct layers based on their position across the various stages of the ML pipeline. Our taxonomy, as illustrated in Figure~\ref{fig:taxonomy}, outlines the following four distinct layers: 
\begin{itemize}[left=0pt]
    \item \hyperref[layer:dp_technique]{\textbf{DP technique}}: The DP algorithms implemented to achieve DP guarantees in ML problems (e.g., DP-SGD).
    \item \hyperref[layer:ml_algorithm_and_hyperparameters]{\textbf{ML algorithm \& hyperparameters}}: The ML algorithm and the training hyperparameters used to train the model.
    \item \hyperref[layer:training_dataset]{\textbf{Training dataset}}: The dataset used to train the ML model.
    \item \hyperref[layer:underlying_distribution]{\textbf{Underlying distribution}}: The underlying distribution from which the training dataset is drawn.
\end{itemize}
Within each layer, we have identified several factors that are likely to influence how the privacy-preserving technique exacerbates disparities. For simplicity, we refer to the four layers as the \emph{DP layer}, the \emph{model layer}, the \emph{dataset layer}, and the \emph{distribution layer} throughout the rest of the paper.

The taxonomy is motivated by the observation that the behavior of an ML model is determined by various components of the ML pipeline rather than by any isolated component. Although existing work in this area focuses predominantly on the role of the DP technique in contributing to the disparities, we observe that factors within other stages of the ML pipeline interact with the DP techniques, compounding their effects, and influencing the extent to which these techniques create or exacerbate disparities.
The taxonomy is complete, as it captures all possible factors within the ML pipeline.
Moreover, the layered structure of the taxonomy conveys causal paths across stages in the ML pipeline: for instance, a factor in the training dataset may propagate up through the stack and exacerbate the effect of the DP technique on the disparities. These causal paths may reveal that the issues stem from lower layers in the taxonomy rather than the top layers, which so far have received the most attention from the research community.

From the perspective of both ML researchers and practitioners, this taxonomy provides useful context about the factors. The layers in the taxonomy assume various levels of control, ranging from the bottom layer, the underlying distribution, which is not typically under the control of ML practitioners, to the top layer, the DP technique, which allows fine-grained control over the operation of the DP mechanism. Therefore, researchers and practitioners can reason about and target mitigation strategies for each layer accordingly.

In this section, we provide a detailed review and analysis of the contributing factors in each layer, as identified in the literature. These factors are categorized according to the established taxonomy, as illustrated in Table~\ref{tab:listfactors}. For each factor, we present supporting evidence of its contribution to the issue, along with findings from Section \ref{sec:causal} on its causal necessity.

\begin{table*}[ht]
\caption{Contributing factors, along with supporting evidence of their contribution to the exacerbation effect within a fairness notion. The last column provides the conclusion of our analysis on the causal necessity of each factor along with the evidence that supports it (if any). We indicate the conclusion with a symbol: $\checkmark$: necessary for all fairness notions; $\RIGHTcircle$: strong evidence for its necessity for the indicated notions; $\times$: unnecessary for all notions; $\otimes$: strong evidence against its necessity for the indicated notions; $\invdiameter$: inconclusive due to lack of evidence. None of the factors was deemed sufficient on its own.}
\centering
\label{tab:listfactors}
\resizebox{\textwidth}{!}{
\begin{NiceTabular}{l|lllc}
    \CodeBefore
        \rectanglecolor{gray!10}{3-2}{3-5}
        \rectanglecolor{gray!10}{5-2}{6-5}
        \rectanglecolor{gray!10}{8-2}{8-5}
        \rectanglecolor{gray!10}{10-2}{10-5}
        \rectanglecolor{gray!10}{12-2}{12-5}
    \Body
\toprule
& & \multicolumn{2}{c}{\textbf{Supporting evidence for each fairness notion (abbr.\ in Section~\ref{fairness_notions})}}  \\
\cmidrule(lr){3-4}
\textbf{Layer}                       & \textbf{Contributing factor}                 & \textbf{Evidence of contribution} &            \textbf{Evidence of conclusion on causal necessity} & \\
\midrule
\multirow{4}{*}{\hyperref[layer:dp_technique]{DP technique}}   & \raisebox{0.5ex}{\hyperlink{factors_within_dpsgd}{Gradient clipping (DP-SGD)}}      & \raisebox{0.5ex}{AP\textsuperscript{\cite{bagdasaryan2019differential, xu2020removing, esipova2022disparate, liu2022mitigating}} RP\textsuperscript{\cite{tran2021differentially,esipova2022disparate}}}  & \raisebox{0.7ex}{$\times$: -}  \\
                                     & \hyperlink{factors_within_dpsgd}{DP noise addition}         & AP\textsuperscript{\cite{bagdasaryan2019differential}} RP\textsuperscript{\cite{tran2021differentially}} & $\checkmark$: - \\
                                     & \multirow{2}{*}{\raisebox{-0.5ex}{\hyperlink{choice_of_dp_algorithm}{Choice of DP algorithm}}}  & \raisebox{-0.5ex}{AP\textsuperscript{\cite{makhlouf2024impact, liu2022mitigating, carey2023randomized, uniyal2021dp, mangold2023differential}} DEP\textsuperscript{\cite{makhlouf2024impact, makhlouf2024systematic, mangold2023differential}} EOD\textsuperscript{\cite{mangold2023differential, noe2022exploring}}}  & \multirow{2}{*}{\raisebox{-0.3ex}{$\times$: -}} & \\
                                     & & \raisebox{-0.5ex}{EOP\textsuperscript{\cite{makhlouf2024impact, mangold2023differential, makhlouf2024systematic}} PEP\textsuperscript{\cite{makhlouf2024impact}} PRP\textsuperscript{\cite{makhlouf2024impact}} RP\textsuperscript{\cite{tran2021differentially, noe2022exploring, xu2019achieving}}} &  & \\
\midrule

\multirow{2}{*}{\hyperref[layer:ml_algorithm_and_hyperparameters]{\makecell[c]{ML algorithm \\ \& hyperparam.}}} & \raisebox{0.5ex}{\hyperlink{model_capacity_and_architecture}{Model capacity and architecture}} & \raisebox{0.5ex}{AP\textsuperscript{\cite{mangold2023differential}} DEP\textsuperscript{\cite{mangold2023differential, de2023empirical}} EOP\textsuperscript{\cite{mangold2023differential}} EOD\textsuperscript{\cite{mangold2023differential, de2023empirical}} PRP\textsuperscript{\cite{de2023empirical}}} & 
\raisebox{0.5ex}{$\otimes$: AP\textsuperscript{\cite{sanyal2022unfair}}} \\
                                                 & \raisebox{-0.5ex}{\hyperlink{training_hyperparameters}{Training hyperparameters}}           & \raisebox{-0.5ex}{DEP\textsuperscript{\cite{de2023empirical}} EOD\textsuperscript{\cite{de2023empirical}} PRP\textsuperscript{\cite{de2023empirical}}} & \raisebox{-0.5ex}{$\otimes$: AP\textsuperscript{\cite{sanyal2022unfair}}}  \\
\midrule
\multirow{3}{*}{\hyperref[layer:training_dataset]{\makecell[c]{Training dataset}}}               & \raisebox{0.5ex}{\hyperlink{dataset_size}{Dataset size}}                    & \raisebox{0.5ex}{AP\textsuperscript{\cite{mangold2023differential, arasteh2023private}} DEP\textsuperscript{\cite{mangold2023differential}} EOP\textsuperscript{\cite{mangold2023differential, cummings2019compatibility}} EOD\textsuperscript{\cite{mangold2023differential}}} & \raisebox{0.5ex}{$\RIGHTcircle$: AP\textsuperscript{\cite{mangold2023differential}} DEP\textsuperscript{\cite{mangold2023differential}} EOP\textsuperscript{\cite{mangold2023differential, cummings2019compatibility}} EOD\textsuperscript{\cite{mangold2023differential}}} \\
                                                & \raisebox{-0.3ex}{\hyperlink{input_norm}{Input norms}}                      & \raisebox{-0.3ex}{RP\textsuperscript{\cite{tran2021differentially}}} & $\invdiameter$: - \\
                                                & \raisebox{-0.5ex}{\hyperlink{data_quality}{Data quality}}                     & \raisebox{-0.5ex}{AP\textsuperscript{\cite{arasteh2023private}}} & \raisebox{-0.5ex}{$\invdiameter$: -}  \\
\midrule
\multirow{2}{*}{\hyperref[layer:underlying_distribution]{\makecell[c]{Underlying \\ distribution}}}       & \raisebox{0.5ex}{\hyperlink{group_imbalance}{Group imbalance}}                 & \raisebox{0.5ex}{AP\textsuperscript{\cite{sanyal2022unfair, farrand2020neither, xu2020removing}}} & \raisebox{0.5ex}{$\otimes$: AP\textsuperscript{\cite{farrand2020neither,esipova2022disparate}}} \\
                                                & \raisebox{-0.7ex}{\hyperlink{group_distance}{\makecell[l]{Group distance to decision boundary}}} & \raisebox{-0.7ex}{AP\textsuperscript{\cite{mangold2023differential}} DEP\textsuperscript{\cite{mangold2023differential}} EOP\textsuperscript{\cite{mangold2023differential}} EOD\textsuperscript{\cite{mangold2023differential}} RP\textsuperscript{\cite{tran2021differentially, xu2019achieving}}}  & \raisebox{-0.7ex}{$\RIGHTcircle$: AP\textsuperscript{\cite{mangold2023differential}} DEP\textsuperscript{\cite{mangold2023differential}} EOP\textsuperscript{\cite{mangold2023differential}} EOD\textsuperscript{\cite{mangold2023differential}} RP\textsuperscript{\cite{tran2021differentially, xu2019achieving}}}& \\
\bottomrule
    \CodeAfter
    \end{NiceTabular}}
\end{table*}

\subsection{DP Technique}
\label{layer:dp_technique}
Since most studies observe an amplification of disparities when applying a DP-based technique to train the model, it is tempting to conclude that the DP technique is the sole cause of the problem.
Thus, rather unsurprisingly, a substantial body of work has focused on analyzing the impact of factors in the top layer of the taxonomy: the DP layer. Given that DP-SGD is the predominant DP technique for PPML, we will first discuss factors related to DP-SGD, and then focus on other, less common, DP-based PPML algorithms.

\para{Factors within DP-SGD.}
\hypertarget{factors_within_dpsgd}
Several studies quantify the disparate impact that DP-SGD can have on model outcomes. Although several components of DP-SGD might play a role in privacy and fairness, the reviewed studies focus only on two major components: \emph{gradient clipping} and \emph{noise addition}.

In the first study that observed the exacerbation effects of DP-SGD, Bagdasaryan et al.\ hypothesize that gradient clipping and noise addition, the two main components of DP-SGD, are both contributing factors~\cite{bagdasaryan2019differential}. They substantiate their argument with empirical evaluations that isolate the effects of each of these two components. Their findings demonstrate that applying DP-SGD on the MNIST dataset~\cite{deng2012mnist} with either only noise addition or only gradient clipping results in a reduced disparate impact in accuracy parity compared to the combination of both. This suggests that both components of DP-SGD are contributing factors to the exacerbation.

In line with this observation, Tran et al.\ conduct a quantitative analysis of how DP-SGD contributes to the exacerbation of disparities~\cite{tran2021differentially}. Their findings also highlight the role of the two components of DP-SGD in contributing to the issue. The authors quantify the group excessive risk within ERM, which measures how DP-SGD alters the loss disparity between different groups. They decompose this group excessive risk into the individual contributions from gradient clipping and noise addition, both of which depend on the Hessian of the loss function. This decomposition also enables them to analytically explore how the properties of the Hessian, the loss gradients, and DP-SGD hyperparameters interact to influence the exacerbation of disparities.

Xu et al.\ are among the first to theoretically analyze how gradient clipping exacerbates existing accuracy disparities~\cite{xu2020removing}. Their empirical investigations, conducted on both vision (MNIST) and census (Adult~\cite{adult_dataset} and Dutch~\cite{vzliobaite2011handling}) datasets, reveal that privacy costs disproportionately affect different groups due to variations in average group gradient norms, which amplifies unfairness. Specifically, underrepresented groups often have larger average gradient norms. In DP-SGD, however, the clipping bound is uniformly applied across all groups, without accounting for this difference. This leads to an uneven privacy-utility trade-off: underrepresented groups or groups with more complex data distributions (e.g., elderly patients exhibiting more intricate and irregular patterns in medical images) experience greater utility loss because their gradients are more severely impacted by clipping.

Esipova et al.\ decompose the error introduced by gradient clipping into two subcomponents: \emph{direction errors} and \emph{magnitude errors}~\cite{esipova2022disparate}. By thresholding the norm of individual gradients, gradient clipping modifies both the magnitude and direction of the aggregated batch gradient~[\S 4.2 in 9]. Based on their analysis, Esipova et al.\ concluded that within the errors caused by gradient clipping, direction errors dominate over magnitude errors in exacerbating accuracy disparity and risk disparity, pointing to gradient \emph{misalignment} as the most significant source of exacerbation of disparities in DP-SGD.

\begin{highlight}
    DP-SGD's \textbf{gradient clipping} mechanism applies a uniform clipping threshold across groups with varying gradient magnitudes, which can induce an uneven effect of gradient clipping and thus amplify existing performance disparities between groups.
\end{highlight}

\para{Choice of DP algorithm.}
\hypertarget{choice_of_dp_algorithm}
Besides DP-SGD, the literature has studied the disparate impact of other DP techniques, evaluating their effect on the exacerbation of disparities compared to DP-SGD. Most of these techniques also add noise to the model parameters, either directly or indirectly, during the training of the model, as explored in works on PATE~\cite{uniyal2021dp}, DP-FedAvg~\cite{liu2022mitigating}, Functional Mechanism~\cite{xu2019achieving}, and Output Perturbation~\cite{mangold2023differential,tran2021differentially}. Other studies investigate DP algorithms that add noise during the data collection and preparation phase, such as Randomized Response \cite{carey2023randomized,makhlouf2024impact,makhlouf2024systematic} and DP-PCA \cite{noe2022exploring}.

Uniayl et al.\ demonstrate that while PATE (Private Aggregation of Teacher Ensembles)~\cite{papernot2016semi} also has a disparate impact on group accuracy, it is significantly lower than that of DP-SGD~\cite{uniyal2021dp} for neural networks on MNIST and SVHN~\cite{netzer2011reading} datasets. However, unlike DP-SGD, PATE follows a semi-supervised approach, assuming access to additional unlabeled data to train a \emph{student} model. This assumption limits its applicability to sensitive domains, such as healthcare, where access to such data is unlikely~\cite{uniyal2021dp}. Notably, to our knowledge, this is the only study that compares the exacerbated impact of DP-SGD with another DP algorithm under the same conditions.

Since the findings of previous work paint a bleak picture of the potential of DP-SGD to achieve equal utility reduction across groups, several studies propose and evaluate alternative DP algorithms. Xu et al.\ achieve a differentially private and fair logistic regression model through two modifications of the objective function~\cite{xu2019achieving}. To achieve privacy, they follow the Functional Mechanism~\cite{zhang2012functional}, an algorithm to ensure DP in logistic regression by injecting the noise into the model's objective function. To achieve fairness, they add a penalty term to the objective function that encourages models with equidistant decision boundaries across groups.
This study demonstrates that at least for logistic regression models, solutions exist that guarantee both DP and empirical risk parity on census datasets like Adult and Dutch. However, these solutions do not apply to highly nonlinear models like deep neural networks. 

In addition to DP algorithms implemented during training, several studies have also analyzed DP algorithms implemented in other phases of the ML pipeline. Carey et al.\ show that applying Randomized Response~\cite{warner1965randomized}, an LDP technique, does not result in a disparate impact on the private model's accuracy across different demographic groups~\cite{carey2023randomized} on various tabular and vision datasets. However, note that in Randomized Response, data owners add noise locally and share the noisy data points with an aggregator that combines them into the final training dataset. Because data owners do not have information on the other training data points, they must be conservative and add noise to cover all possible cases, which in most practical scenarios destroys the utility of the resulting model. Therefore, Randomized Response can ensure reasonable utility only in settings with large numbers of data contributors~\cite{juarez2023you}. 

Following this work, Makhlouf et al.\ demonstrate that implementing multi-dimensional, rather than single-dimensional, LDP is an even more effective strategy to mitigate potential disparities for multiple fairness notions~\cite{makhlouf2024impact}. They first show that applying LDP with Randomized Response reduces disparity, confirming the findings of Carey et al.~\cite{carey2023randomized}. Furthermore, they illustrate that protecting multiple sensitive attributes leads to a more significant disparity reduction than a single attribute. Makhlouf et al.\ also show that applying LDP by adding noise to sensitive attributes could enhance demographic parity and equal opportunity for unprivileged groups~\cite{makhlouf2024systematic}. 

As promising as these results may appear, it remains uncertain whether DP during pre-processing offers the same privacy protection as during training. Noe et al.\ show that this may not be the case. When they apply DP to PCA (Principal Component Analysis~\cite{abdi2010principal}) for the CivilComments~\cite{pmlr-v139-koh21a} dataset, DP does not significantly affect equalized odds or empirical risk parity of the trained model \cite{noe2022exploring}. However, upon closer examination, it is noted that this approach may not offer as robust a privacy guarantee. The authors illustrate this with an example of a successful attack against models whose features are selected through DP-PCA, suggesting that privacy properties may not be guaranteed in the downstream classification task. Additionally, it is acknowledged that LDP at the dataset level generally impacts model utility more than GDP techniques like DP-SGD, limiting its use in practice~\cite{ponomareva2023dp}.

Output Perturbation is another technique that ensures DP within the supervised learning pipeline, where noise calibrated to the sensitivity of the training method is added directly to the final learned model parameters from training~\cite{chaudhuri2011differentially}.
Mangold et al.\ investigate Output Perturbation along with DP-SGD, and their results show no significant differences between Output Perturbation and DP-SGD on both vision (CelebA~\cite{liu2015faceattributes}) and tabular (folktables~\cite{ding2021retiring}) datasets in terms of AP, DEP, EOP, and EOD~\cite{mangold2023differential}, suggesting that both DP mechanisms produce similar disparities across these fairness notions.
Tran et al. show that Output Perturbation can introduce disparities in excessive risk when groups have different local curvatures of their loss and that the magnitude of the disparities in excessive risk is proportional to the amount of added noise~\cite{tran2021differentially}.
The authors state that while Output Perturbation does not guarantee fairness, normalizing the input values for each group independently could lead to it. However, implementing this solution requires access to sensitive group attributes at inference time, which is a strong requirement, as they are often unavailable due to privacy and legal constraints.
In addition, the applicability of Output Perturbation is limited compared to DP-SGD in the industry because it restricts the number of queries that can be made, and it can only guarantee DP on twice-differentiable and convex loss functions~\cite{tran2021differentially}.


\begin{highlight}
    Different \textbf{DP techniques} can lead to varying levels of disparate impact. However, DP techniques exhibiting lower disparate impact may have reduced practical applicability and offer weaker privacy guarantees compared to those with higher disparate impact.
\end{highlight}

\subsection{ML Algorithm \& Hyperparameters}
\label{layer:ml_algorithm_and_hyperparameters}

Though less explored than other layers of the taxonomy, several studies investigate the impact of the learning algorithm and its hyperparameters on the disparate effect of DP.

\para{Model capacity and architecture.}
\hypertarget{model_capacity_and_architecture}
Cheng et al.\ introduce DPNAS, a Neural Architecture Search (NAS) technique optimized for models trained with DP-SGD~\cite{cheng2022dpnas}. This method effectively balances the privacy-utility trade-off, outperforming SoTA models on benchmarks like MNIST and CIFAR-10~\cite{Krizhevsky2009}. The analysis of DPNAS-generated architectures also sheds light on design strategies for high-utility DP-SGD-trained models. DPNAS does not necessarily reduce disparate impacts, as improvements in overall utility may not be uniformly distributed across different groups. However, extending on this work, de Oliveira et al. show that compared to a baseline DP-SGD model---where \emph{model architecture} and hyperparameters were tuned for overall accuracy before applying DP-SGD---a search over architectures and hyperparameters specifically with DP-SGD can yield models with less disparate impact. These models outperform the baseline in fairness measures like DEP, EOD, and PRP on census datasets~\cite{de2023empirical}.

Mangold et al.\ provide a theoretical analysis focused on DP-SGD and Output Perturbation, proving high probability bounds \hypertarget{prob_bound} on fairness reduction based on the model's confidence margin. The bound shows that the disparity (in AP, DEP, EOP, and EOD) induced by DP decreases at a rate of $\mathcal O(\sqrt{p/n})$, where \(p\) denotes model size and \(n\) represents dataset size. This result also indicates that \emph{model capacity} is a factor in the exacerbation of disparities. Although empirical validation is lacking, the theoretical results suggest that, for a fixed dataset size, simpler models are expected to attenuate the disparities.

\para{Training hyperparameters.}
\hypertarget{training_hyperparameters}
Yao et al.\ highlight that \emph{training hyperparameters}, such as batch size and learning rate, play a pivotal role in shaping the Hessian spectrum of a classifier~\cite{yao2018hessian}. This implies that training hyperparameters can have an impact on the gradient clipping mechanism of DP-SGD. These insights suggest that various key hyperparameters, including batch size and learning rate, are factors that influence the disparities caused by applying DP to ML training.

The study by de Oliveira et al.\ supports this hypothesis~\cite{de2023empirical}. The authors show that models trained by constraining hyperparameter tuning with DP-SGD can exhibit reduced disparities across various fairness notions compared to the baseline DP-SGD model, as explained before. Although this work does not provide a detailed analysis of the individual contribution of each hyperparameter, it provides evidence of the contribution of training hyperparameters to the unfairness exacerbation.

\begin{highlight}
    \textbf{Model capacity and architecture}, and \textbf{training hyperparameters} of a model, can affect the exacerbation of disparities when the model is trained with DP. However, the underlying mechanisms remain unclear.
\end{highlight}

\subsection{Training Dataset}
\label{layer:training_dataset}

In practice, the sampling of the training dataset may introduce biases and noise, and thus the distribution of the training data may deviate from the underlying distribution. As a result, the model's ability to generalize, and, in particular, differences in its generalization ability across groups, may be impacted by this sampling bias. Several properties of the training dataset have been identified in the literature as factors that contribute to the exacerbation of such group disparities caused by DP. 

\para{Dataset size.}
\hypertarget{dataset_size}
Among the factors considered in this layer of the taxonomy, \emph{dataset size} has been extensively discussed in the literature and is considered to play a critical role in the disparate effect of DP. Intuitively, this is because the sensitivity of functions related to calculating DP noise is highly dependent on the size of the sample. With a larger sample, the impact of a change in any single data point diminishes, resulting in a lower overall sensitivity. Consequently, less DP noise is required to achieve the same privacy guarantees, thus attenuating the effects of DP.
Mangold et al.\ provide mathematical proof for this intuition~\cite{mangold2023differential}. In the \hyperlink{prob_bound}{probability bound} mentioned before, dataset size is an inversely proportional factor, indicating that, for a fixed model size $p$, the disparity of the model converges to that of its non-private counterpart as dataset size increases.

Cummings et al.\ theoretically prove the satisfiability of $(\varepsilon, \delta)$-DP and \emph{approximate fairness} for a PAC learner~\cite{cummings2019compatibility}, reconciling DP and fairness. Their main compatibility theorem establishes that, under the assumption of a lower bound on the dataset size, it is possible to simultaneously satisfy approximate fairness in terms of equal opportunity and $(\varepsilon, \delta)$-DP. This result emphasizes the critical role of dataset size as a factor in our taxonomy.

In addition, Arasteh et al.\ provide indirect but valuable empirical evidence that dataset size can be a contributing factor to the exacerbation of accuracy disparities introduced by DP~\cite{arasteh2023private}. In this study, the authors train a model with DP-SGD on a large-scale medical image dataset comprising nearly $200,000$ samples. This private model does not exhibit significant disparate impact compared to non-private models. The authors attribute the fairness achieved by their model partly to the large size of their dataset, supporting previous claims on the attenuation effect of having large dataset sizes in the learning problem at hand.

\para{Input norms.}
\hypertarget{input_norm}
Another intriguing observation is that variations in the norm of input values among different groups can also exacerbate unfairness under DP~\cite{tran2021differentially}. Tran et al.\ show that the \emph{input norms} of a population group in the training dataset influence the gradients and Hessian of the loss function. Notably, they find that in a linear model with cross-entropy loss for a multi-class classifier, the gradient norms are proportional to the input norms. Consequently, due to the gradient clipping mechanism, the input norms of a group in the data act as an indirect factor that contributes to the excessive risk within that group. The authors thus identify differences in input norms among different groups as a proxy for disparity: groups with larger input norms are likely to experience greater disproportionate impacts during private training compared to groups with smaller input norms.

\para{Data quality.}
\hypertarget{data_quality}
Another relevant insight from the literature is that \emph{quality} of the training data (e.g., image resolution) could also mediate in the disparate effect of DP~\cite{arasteh2023private}. For instance, Arateh et al.\ propose a learning algorithm that preserves group parities while providing strong DP guarantees, and attribute this optimistic result partly to the higher image quality of their medical image dataset compared to the datasets used by studies reporting disparities~\cite{arasteh2023private}. However, the authors do not provide sufficient experimental validation to support this hypothesis.


\begin{highlight}
    Various training dataset properties have been shown to influence the disparate effect of DP, among which \textbf{training dataset size} is the most prominent. Both theoretical and empirical analysis demonstrate the attenuation effect of large datasets on disparate impacts.
\end{highlight}

\subsection{Underlying Distribution}
\label{layer:underlying_distribution}
We have identified two factors that fall into the bottom layer of our taxonomy, corresponding to properties of the underlying distribution: \emph{group imbalance} and \emph{group distance to the decision boundary}. A group imbalance occurs when there is a significant difference between the marginal probabilities of the groups, with the likelihood of encountering individuals of one group being significantly higher than the other. Although group imbalance can also originate at the dataset layer, prior work has only studied the impact of distributional group imbalance. Group distance to the decision boundary refers to the relative distance of the groups to the model's decision boundary. Previous work suggests that these factors play a significant role in creating or amplifying disparate model outcomes when DP is integrated into the ML pipeline.

\para{Group imbalance.}
\hypertarget{group_imbalance}
Bagdasaryan et al.\ emphasize that DP-SGD exacerbates disparities against minority groups~\cite{bagdasaryan2019differential}. They observe that group size critically affects fairness, with smaller groups experiencing greater utility loss due to imbalance.

Xu et al.\ offer an explanation for why minority groups can suffer an increased accuracy loss when DP is implemented~\cite{xu2020removing}. The authors show that the imbalance in group size is a primary factor that influences the magnitude of the gradient norm of the group: minority groups have smaller group sizes, resulting in larger average gradient norms during training. As a result, they suffer from larger disparities due to the uniform treatment of DP-SGD's gradient clipping across all groups.

With the aim of analyzing the effect of the group marginal distributions on the disparate effect of DP, Sanyal et al.~\cite{sanyal2022unfair} introduce a model-agnostic theoretical framework that allows the authors to prove an impossibility result: in the case of \emph{long-tailed} group distributions (indicating a significant imbalance across majority and minority groups), it is not possible to have accurate ML algorithms that are private and also maintain accuracy for minority groups. The authors support their theoretical findings with experimental evidence on synthetic and vision datasets such as CelebA. In their experiments, the authors consider minority groups with only a few samples each, creating an extreme sample group size imbalance.

However, some studies challenge the importance of group imbalance in contributing to the exacerbation issue of DP-SGD. Farrand et al.\ investigate how different degrees of group imbalance affect the accuracy disparity in models trained with DP-SGD~\cite{farrand2020neither}. The authors demonstrate that even slight group imbalances can lead to disparities. 
Additionally, for smile detection tasks, accuracy disparities between groups only increase significantly at extreme group imbalance (a majority-to-minority group ratio of 999:1), potentially indicating a weaker link between group imbalance and DP-induced disparities compared to other factors in our taxonomy. Esipova et al.\ provide additional evidence, in which they experiment with an artificially balanced version of the Adult dataset regarding sex (14,000 samples in each group); despite the perfect balance, the authors still observe that the accuracy disparity between the two groups is exacerbated by DP-SGD~\cite{esipova2022disparate}.

\begin{highlight}
    The impact of \textbf{group imbalance} remains contested. While several studies show that group imbalance plays a significant role, others have noted that disparities can still be exacerbated without group imbalance.
\end{highlight}

\para{Group distance to the decision boundary.}
\hypertarget{group_distance}
We categorize this factor under the distribution layer of the taxonomy rather than the model layer because although the group distance to the decision boundary is indeed determined by both the data distribution and the model, the underlying distribution is more decisive: practitioners may be able to modify the decision boundary, but cannot change the underlying data distribution.

Several studies have identified the \emph{group distance to the decision boundary} as a key factor contributing to the exacerbation issues associated with DP. Tran et al.\ are among the first to make this observation~\cite{tran2021differentially}. Their analysis reveal that the primary factor influencing the impact of noise addition in DP-SGD on exacerbating excessive risk disparity is the Hessian of the loss, which, in turn, is determined by the group's distance to the decision boundary. This observation establishes group distance to the decision boundary as an indirect underlying factor in the disparate impacts of DP-SGD.

Mangold et al.\ reach similar conclusions~\cite{mangold2023differential}. They prove that some group fairness metrics (AP, DEP, EOP, and EOD) are pointwise Lipschitz with respect to the model. The pointwise Lipschitz constant explicitly depends on the confidence margin of the model and can be computed from a finite data sample. Their findings underscore the significance of the confidence margin of models in the differential impact of DP; specifically, if the non-private model exhibits high overall confidence (i.e., data points are far from the model's decision boundary for both groups), then DP is less likely to create or amplify disparities.

Furthermore, Xu et al.\ penalize biased decision boundaries by incorporating the distance to the decision boundary as a regularization term in the objective function of logistic regression, favoring models with decision boundaries that have similar distances across group centroids~\cite{xu2019achieving}. The ability of the resulting models to achieve strong DP guarantees without significant demographic disparities is a strong indicator that group distance to the decision boundary is essential in mitigating the exacerbation issues associated with DP.

\begin{highlight}
\textbf{Group distance to the decision boundary} influences the disparity caused by DP. While not explicitly studied, this effect is intuitive: DP noise shifts the decision boundary, and groups closer to it are more affected, resulting in more errors and reduced fairness.
\end{highlight}

\section{Causal Analysis of Factors}
\label{sec:causal}

Upon reviewing all contributing factors in our taxonomy, we come to question their necessity and sufficiency. A factor is necessary if it is required for the occurrence of the phenomenon, and is sufficient if it guarantees it.
Identifying the necessity and sufficiency of a factor helps us to understand the causal relationship it has with the phenomenon and informs recommendations for mitigation.
This causal analysis is based on evidence found in the reviewed studies.
Table~\ref{tab:listfactors} summarizes the factors, the supporting evidence for their necessity or/and sufficiency for a given set of fairness notions, and our conclusions regarding their causal links to the issue.

\subsection{Causal Analysis}
We first discuss the necessity of the factors and then draw conclusions on their sufficiency.

\para{DP technique.}
The causal analysis of this layer relies on logical arguments rather than empirical evidence, as most of the factors in this layer can be ruled necessary or unnecessary based on their definitions.

The only factor that we deem necessary ($\checkmark$) for the exacerbation effect is the \emph{noise addition} operation inherent to the DP mechanism.
The rationale is that noise addition is the fundamental strategy through which DP is achieved, which necessarily perturbs the model's decision boundary. If the decision boundary was not perturbed, there would be no exacerbation effect. Therefore, adding noise and thus perturbing the decision boundary is a necessary condition for DP's exacerbation effect.

On the other hand, we consider the other factors in this layer unnecessary ($\times$). The argument is again based on the observation that perturbing the decision boundary (through noise addition) is, by definition, inherent to DP, irrespective of the specific DP technique, its implementation or configuration (e.g., DP-SGD's gradient clipping); therefore, the disparities can arise regardless of these choices. Although gradient clipping can significantly exacerbate unfairness, it is not necessary, as other DP methods without it can also lead to unfairness.  

In conclusion, within this layer we identify noise addition as the only necessary factor for DP's exacerbation effect rather than the choice of the DP mechanism or its configuration. 



\para{ML algorithm \& hyperparameters.}
According to our analysis, none of the factors within this layer is necessary for the exacerbation issues of DP.

This conclusion is supported by the following argument: consider a training dataset with unfavorable characteristics for achieving fairness; in such cases, regardless of the choice of model or hyperparameters, exacerbation issues still exist. For instance, Sanyal et al.\ provide such evidence by demonstrating that for long-tail distributions, achieving both privacy and equal utility reduction among groups while preserving reasonable accuracy is not possible, regardless of the choice of the learning algorithm or architecture~\cite{sanyal2022unfair}. This indicates that the contributing factors within the model layer are not necessary for the occurrence of exacerbation issues ($\otimes$).

\para{Training dataset.} From our analysis, we conclude that a \emph{small} dataset size is likely to be a necessary condition, while other factors within this layer are unnecessary for the disparity exacerbation of DP.

The rationale is that the sensitivity of the learning algorithm is inversely proportional to the dataset size. If the training dataset is larger, the contribution of any single data point is smaller, resulting in reduced sensitivity. Therefore, less noise is needed to achieve the same DP guarantees, leading to milder changes in utility. When the dataset size is sufficiently large, the noise is not sufficiently large to have an impact on utility, and consequently on the difference in utility across groups.

This intuition is strongly supported ($\RIGHTcircle$) by both empirical~\cite{mangold2023differential} and theoretical~\cite{cummings2019compatibility} studies. Mangold et al.\ observe that the disparity diminishes as dataset size increases~\cite{mangold2023differential}. In line with this observation, Cummings et al.\ establish that surpassing a certain threshold in dataset size enables both privacy and approximate fairness in ML models~\cite{cummings2019compatibility}. Therefore, we conclude that a dataset size smaller than a certain threshold is necessary for the exacerbation issue to arise.

We classify \emph{input norms} and \emph{data quality} as not necessary for the exacerbation of disparities, mainly due to the lack of support in the literature ($\invdiameter$). The only study identifying input norms as a potential factor suggested its effect is indirect and did not provide empirical validation regarding its direct impact on exacerbating disparities~\cite{tran2021differentially}. Similarly, the only study suggesting that data quality might be a relevant factor lacks empirical evidence to support a direct causal link between data quality and disparity exacerbation~\cite{arasteh2023private}. 

\para{Underlying distribution.} Based on our analysis, we conclude that \emph{group imbalance} is not necessary, and that \emph{group distance to the decision boundary} is likely necessary for the exacerbation issue associated with DP in ML.

We consider \emph{group imbalance} not to be a necessary condition for exacerbating disparities. Research has shown that in certain settings only extreme imbalances significantly worsen the exacerbation~\cite{farrand2020neither}. Moreover, these studies indicate that a slightly imbalanced or even perfectly balanced dataset can still experience unfairness exacerbation by DP~\cite{farrand2020neither,esipova2022disparate}, strongly suggesting that group imbalance is not necessary to occurrence of the issue and is dominated by other contributing factors in our taxonomy ($\otimes$).

We consider \emph{group distance to the decision boundary} to be necessary for the exacerbation issue. This is grounded in the intuition that if distances to the decision boundary for different groups are equal, then perturbing the decision boundary by DP noise would have similar effects on the groups, and is thus unlikely to have disparate effects. Hence, there has to be a difference in the distances of each group to the decision boundary for the issue to arise. This intuition is supported by multiple pieces of evidence ($\RIGHTcircle$) discussed in Section~\ref{layer:underlying_distribution}.

\subsection{Conclusion}

We conclude that besides DP's \emph{noise addition} ($\checkmark$) as an intrinsic direct cause, \emph{dataset size} ($\RIGHTcircle$) and \emph{group distance to the decision boundary} ($\RIGHTcircle$) are the only other factors in our taxonomy that are likely necessary for the exacerbation issue associated with implementing DP in ML. This also implies that the presence of these two factors simultaneously is likely to be sufficient for the issue to manifest. Since multiple factors are considered possibly necessary for the issue, it follows that \emph{no single factor alone is sufficient}, thus concluding our causal analysis given the existing studies in the field.

Contrary to the common intuition that factors regarding the DP technique should be the most responsible for the issue, we find that they are neither necessary nor sufficient, except for the addition of DP noise, which is inherent to the design of DP. Instead, we find that the factors that are more likely to be necessary for the issue lie in the lower layers of our taxonomy: the dataset layer and the distribution layer, prompting future research to focus on the interplay of the DP technique with these underlying factors.

The four layers of our taxonomy are structured in a hierarchy, and there is a tight connection between each adjacent pair of layers: as shown in Figure~\ref{fig:taxonomy}, the training dataset is drawn from the underlying distribution, ML algorithm \& hyperparameters are chosen based on the characteristics of the training dataset, and the choice of DP technique is determined by both the chosen ML algorithm \& hyperparameters, as well as the training dataset. Thus, the distribution and the dataset layer are the most fundamental of all, and thus their roles are more profound---albeit also more indirect---in any issue that might arise in the predictive behavior of the models, including the exacerbation of performance disparities by DP.

\section{Mitigation of Unfairness}
This section provides an overview of strategies proposed in recent years to mitigate the exacerbation issues of DP in ML. Given that these mitigation strategies often target specific contributing factors, we categorized these strategies also based on our taxonomy of contributing factors, as shown in Table~\ref{tab:mitigation}.


\begin{table*}[ht]
\caption{Mitigation strategies categorized by the taxonomy.}
\begin{center}
\begin{NiceTabular}{l|l|l}
    \CodeBefore
        \rectanglecolor{gray!10}{2-2}{2-3} 
        \rectanglecolor{gray!10}{4-2}{4-3} 
        \rectanglecolor{gray!10}{6-2}{6-3} 
        \rectanglecolor{gray!10}{8-2}{8-3} 
        \rectanglecolor{gray!10}{10-2}{10-3} 
        \rectanglecolor{gray!10}{12-3}{12-3} 
    \Body
    \toprule
    \textbf{Layer} & \textbf{Ref.} & \textbf{Mitigation Strategy} \\ \midrule
    \multirow{6}{*}{DP technique} & \raisebox{0.3ex}{\cite{xu2020removing}} & \raisebox{0.2ex}{DPSGD-F: Set different clipping thresholds for different groups} \\ 
                                  & \raisebox{-0.3ex}{\cite{liu2022mitigating}} & \raisebox{-0.3ex}{Adjust instance influence dynamically with self-adaptive DP mechanism} \\ 
                                  & \raisebox{-0.3ex}{\cite{tran2023fairdp}} & \raisebox{-0.3ex}{FairDP: Train separate models for separate groups} \\ 
                                  & \raisebox{-0.3ex}{\cite{esipova2022disparate}} & \raisebox{-0.3ex}{DPSGD-Global-Adapt: Align the direction of clipped gradients} \\ 
                                  & \raisebox{-0.3ex}{\cite{uniyal2021dp}} & \raisebox{-0.3ex}{Use PATE (an alternative DP algorithm) instead of DP-SGD} \\ 
                                  & \raisebox{-0.5ex}{\cite{tran2021differentially}} & \raisebox{-0.5ex}{Equalize terms contributing to excessive risk across all groups} \\ \midrule
    \multirow{2}{*}{\begin{tabular}[c]{@{}l@{}}ML algorithm\\ \& hyperparameters\end{tabular}} & \raisebox{0.3ex}{\cite{de2023empirical}} & \raisebox{0.2ex}{Conduct a model architecture and hyperparameter search} \\ 
                                  & \raisebox{-0.5ex}{\cite{cummings2019compatibility}} & \raisebox{-0.5ex}{Design a PAC learner considering both fairness and privacy} \\ \midrule
    Training dataset              & \cite{srivastava2024amplifying} & Apply Counterfactual Data Augmentation technique \\ \midrule
    Underlying distribution        & \cite{xu2019achieving} & Implement decision boundary fairness constraint\\ \bottomrule
\end{NiceTabular}
\label{tab:mitigation}
\end{center}
\end{table*}

\subsection{Mitigation Strategies}
\label{sec:mitigation_strategies}

\para{DP technique.}
Many mitigation strategies focus on optimizing the gradient clipping mechanism in DP-SGD. Xu et al.\ are among the first to propose such mitigation strategies \cite{xu2020removing}. The authors design an algorithm called DPSGD-F, a variant of DP-SGD for removing disparate impact caused by DP-SGD and reducing accuracy disparity. DPSGD-F adaptively sets a different gradient clipping threshold for each group. The method is based on the observation that negatively impacted groups tend to have large gradient norms which are, in turn, more affected by the clipping. Therefore, in DPSGD-F, the clipping threshold is adaptively increased for groups with larger gradient norms, based on an estimate of how many gradients per group are larger than a clipping threshold. This estimate can be obtained with DP guarantees without significantly increasing the overall privacy budget. Liu et al.\ conduct a similar study in which they adjust DP-SGD and quantify clipping bias and noise variance to dynamically adjust the sample clipping threshold, resulting in more equitable models~\cite{liu2022mitigating}. However, the DP guarantee of this method is unclear.

Tran et al.\ also introduce a mitigation strategy based on the concept of group-specific clipping~\cite{tran2023fairdp}. This method involves training separate models for individual demographic groups and incorporating group-specific clipping terms for each model to evaluate and bound the disparate impacts of DP. As training progresses, the mechanism aggregates information from each group model and regulates the injection of noise and group-specific clipping to mitigate DP's disparate impacts.

Esipova et al.\ show that gradient misalignment, characterized by directional errors produced by the gradient clipping mechanism, is a dominant factor in DP-SGD contributing to unfairness exacerbation~ \cite{esipova2022disparate}. Building upon this insight, they introduce a mitigation strategy named DPSGD-Global-Adapt, which aims to minimize gradient misalignment. DPSGD-Global-Adapt is a variant of DPSGD-Global \cite{bu2021convergence}, an approach that uniformly scales down all per-sample gradients within a batch to approximately realign the gradients. DPSGD-Global operates by discarding gradients exceeding a certain threshold $Z$ and uniformly scaling down the remaining gradients. While this ensures exact alignment when $Z$ is sufficiently large, a large $Z$ can result in gradients being scaled down too aggressively, leading to slow convergence. On the other hand, a small $Z$ introduces disparities because it discards minority gradients more often. In response to these limitations, DPSGD-Global-Adapt refines DPSGD-Global by first clipping large gradients to $Z$ instead of discarding them, and then adaptively updating $Z$ in each batch. The authors show that DPSGD-Global-Adapt achieves improved fairness in terms of accuracy parity and risk parity compared to DP-SGD.

Tran et al.\ propose a mitigation strategy for risk disparity in DP-SGD, which complements existing work focused primarily on gradient clipping~\cite{tran2021differentially}. Their approach quantifies the contributions of the two main components in DP-SGD, gradient clipping and noise addition, denoted $R^{clip}$ and $R^{noise}$, respectively, toward the overall risk disparity. The authors analyze how these contributions are affected by factors such as the Hessian of the loss, gradients, privacy parameters, and hyperparameters like the clipping bounds. To address these disparities, they introduce a strategy aimed at balancing the contributions of $R^{clip}$ and $R^{noise}$ across all demographic groups during private model training. This is achieved by computing the Hessian of the loss and gradients for each sample at each iteration, enabling more precise control over the model's risk across different groups.

Besides DP-SGD, the works \cite{uniyal2021dp,xu2019achieving,carey2023randomized,makhlouf2024impact,makhlouf2024systematic,noe2022exploring} provide evidence that choosing an alternative DP technique rather than DP-SGD may be an effective mitigation approach. Among these works, Uniyal et al.\ provide the most direct evidence by showing that PATE renders a fairer model in accuracy parity than DP-SGD under the same setting on vision datasets\cite{uniyal2021dp}.

\para{ML algorithm \& hyperparameters.}
As mentioned in Section~\ref{sec:factors}, de Oliveira et al.\ empirically demonstrate that, through an exhaustive search of model architecture and training hyperparameters for DP-SGD, practitioners can reduce disparity exacerbation~\cite{de2023empirical}. Their experiments account for the privacy budget spent during the randomized hyperparameter search by factoring it into the total budget allocated for training the model, thus ensuring a fair comparison between DP models and baseline models.
Their findings indicate that the exacerbation of disparities of DP can be mitigated to some extent by adjusting model configurations. 

Furthermore, Cummings et al.\ propose an alternative mitigation strategy based on the theoretical design of an ML algorithm that is both private and fair \cite{cummings2019compatibility}. Their theoretical analysis offers an efficient PAC learner algorithm that maintains utility while satisfying both privacy and approximate fairness (equal opportunity) with high probability. 

\para{Training dataset.}
Arasteh et al.'s experiments provide support for the hypothesis that ensuring a training dataset of high quality and large size mitigates DP's accuracy disparities~\cite{arasteh2023private}. In their investigation, the authors evaluate a model trained with DP-SGD on a large-scale clinical dataset with high image quality. Their results show that the model exhibits a reasonable utility-fairness compromise. The authors attribute these results to the dataset's scale and quality and suggest that pretraining the models on public datasets is an inexpensive method to further improve fairness outcomes.

Srivastava et al.\ find that disparities in gradient convergence across different groups caused by the clipping may contribute to the exacerbation of unfairness in language models, as reflected in bias metrics such as toxicity \cite{srivastava2024amplifying}. Through an analysis focused on binary gender bias, they illustrate that Counterfactual Data Augmentation (CDA), a recognized method for mitigating bias in language models, can also mitigate disparity exacerbation when DP is involved. Consequently, the combined application of DP and CDA offers a promising solution for fine-tuning language models while preserving both fairness and privacy. Despite studying a generative model, this work can provide insights for classification problems.

\para{Underlying distribution.} Xu et al.\ show that combining Functional Mechanism and decision boundary fairness can mitigate the exacerbation of empirical risk disparity by DP in ERM \cite{xu2019achieving}. The authors introduce a penalty term to the objective function to encourage establishing a decision boundary that reduces disparities in group distance. Given that logistic regression has an analytic expression of the decision boundary, the authors can calculate the exact penalty term encoding group distances, thus accomplishing this mitigation strategy.

\subsection{Limitations}

Despite the promising results of existing mitigation techniques, these approaches are subject to certain limitations that restrict their practicality. Next, we discuss these limitations.

\para{Group label disclosure.} Many mitigation strategies that optimize DP-SGD require group label information for all the data points~\cite{xu2020removing,liu2022mitigating,tran2021differentially,tran2023fairdp}. It appears intuitive that addressing group fairness necessitates designing approaches based on the grouping of the data. However, this information may not always be accessible, as membership in certain groups can be sensitive, with users often preferring to keep it confidential.
Furthermore, access to group membership information is often regulated~\cite{veale2017fairer}. For example, the EU's GDPR mandates that protected attributes, such as gender or race, be collected only under appropriate privacy protections and explicit informed consent. Therefore, the industry tends to be conservative in the collection and utilization of these attributes, particularly in sectors subject to greater scrutiny \cite{bogen2020awareness}.

\para{Repeated data querying.} Some mitigation strategies require repeated querying of the training samples. For instance, in the approach presented by de Oilveira et al., an exhaustive search for optimal model architectures and training hyperparameters can find better fairness trade-offs~\cite{de2023empirical}. This search involves training multiple times on the training dataset which would require adjusting the privacy budget of DP-SGD to account for the additional queries. Due to the composability properties of DP, the privacy budget required by these mitigations can be calculated easily; however, the additional noise required to achieve the same level of privacy may have an impact on the final fairness.
This is a chicken-and-egg problem:
Although stricter DP guarantees can be enforced, doing so may undermine the effectiveness of mitigation techniques and, conversely, relaxing these guarantees increases the privacy risk. Thus, the additional privacy risks associated with repeated data access must be carefully managed within the DP mechanism to maintain a formal DP guarantee while taking into account its impact on the final fairness outcome.

\para{Increased computational costs.} In addition to privacy considerations, another significant concern regarding mitigation is the additional computational costs it entails. For instance, the computation of Hessians at each training step required by Tran et al.'s mitigation strategy is computationally intensive~\cite{tran2021differentially}. Similarly, the model architecture and hyperparameter search strategy outlined in the study of de Oliveira et al.\ also increases computational costs by multiple folds depending on the design, as the search process requires multiple training runs to find the optimal configuration for fairness~\cite{de2023empirical}. While accepting a small added cost is reasonable if the countermeasure is effective, such large costs can discourage the adoption of the proposed mitigation techniques. 


\section{Discussion \& Open Questions}
\label{sec:discussion}

Finally, we present some open questions and research directions based on the gaps identified in the existing literature.

\subsection{Limitations of the Current Causal Analysis}

We identified only 20 works that fall within the scope of this survey, indicating that this research area is still maturing. In particular, the analysis of contributing factors is limited, particularly with regard to determining their relative dominance and inferring interactions between them, given the scarcity of available literature and their diverse experimental settings.

In our causal analysis, we conclude that two factors---\emph{dataset size} and \emph{group distance to the decision boundary}---are especially significant. These factors are likely the most dominant in influencing fairness in PPML. We found that the current literature lacks evidence and analysis on the relative importance of contributing factors, through ablation studies or scaled analyses, which hinders a deeper understanding of the issue and the development of effective mitigation strategies.

Additionally, the interaction between related factors across different layers of the taxonomy is underexplored. For instance, while batch size (model layer) and dataset size (dataset layer) are closely linked, their combined effects remain unstudied. Tran et al.~\cite{tran2021differentially} are among the few to examine such inter-factorial relationships, showing how \emph{group distance to the decision boundary} and \emph{input norms} influence gradients, which in turn affect \emph{gradient clipping} in DP-SGD. Other studies have explored gradient clipping's impact across tasks and datasets~\cite{bagdasaryan2019differential,xu2020removing,esipova2022disparate,liu2022mitigating}, suggesting a connection between \emph{dataset characteristics} and \emph{gradient clipping}, but did not explicitly control for these characteristics when evaluating their interaction with DP-SGD, leaving a gap for future research. 

\subsection{Dataset and Distribution Layers Require More Attention}

Our causal analysis indicates that the most significant factors reside in the lower layers, suggesting that interventions at these levels may have a greater impact. Most existing works on the factorial analyses reviewed in Section \ref{sec:factors} and the mitigation strategies discussed in Section \ref{sec:mitigation_strategies} focus on the DP technique layer. While this focus is justifiable---since it is the introduction of DP that exacerbates existing disparities---it risks overlooking the potential compounded benefits of addressing fairness at the lower layers of the taxonomy. 

We encourage future research to focus on these lower layers, as they are crucial in determining fairness outcomes. This underscores the sociotechnical nature of algorithmic bias, extending beyond DP to broader ML systems, where societal biases skew distributions, resulting in biased datasets~\cite{crawford2017,chen2018my}. Further investigation is needed into the role of distributional properties, individual data characteristics, and group distance to the decision boundary.

At the distribution level, Sanyal et al.\ demonstrate that datasets with long-tailed distributions prevent PPML models from achieving both privacy and group fairness while maintaining high accuracy~\cite{sanyal2022unfair}. Their findings are based on specific distributional assumptions, yet many real-world datasets can present other distributional patterns. This raises an important question: What other distributional properties impact fairness? This is an area of research that remains underexplored and represents a promising direction for future work.

In addition to distributional considerations, we still have a limited understanding of how individuals are impacted by DP’s disparate effects. Addressing this gap could involve leveraging ML explainability techniques~\cite{linardatos2020explainable} to analyze how specific individual characteristics influence prediction shifts under DP. Such an analysis would enable more fine-grained mitigation strategies, focusing on the most vulnerable individuals.

Furthermore, disparities in group distances from decision boundaries is clearly another significant contributing factor that requires more attention. Although several studies have provided strong evidence suggesting that DP noise can perturb decision boundaries in ways that disproportionately impact certain groups~\cite{xu2019achieving,tran2021differentially,mangold2023differential}, more research is needed to theoretically or empirically validate this effect. A better understanding of the underlying mechanisms by which DP noise affects the decision boundaries of a model could pave the way for developing fairer DP algorithms.

\subsection{The Challenges of Conflicting Fairness Definitions}

Research papers often adopt varying fairness notions, making it difficult to compare their conclusions. To rectify this, new studies should clearly justify the choice of specific fairness notions. 
For example, while it is valid to justify focusing on demographic parity if the study is specific to college admissions, studies that focus on investigating the disparate effect of DP must include multiple notions of fairness wherever possible to gain a more comprehensive understanding and facilitate comparison across studies. It is necessary not only for evaluating DP's impact on fairness in ML but also for fairness research in general. This is particularly important given the incompatibility theorem on fairness notions~\cite{miconi2017impossibility,saravanakumar2020impossibility}, which shows that no imperfect model can satisfy all fairness criteria simultaneously. Therefore, comparing conflicting definitions is necessary. In addition, we identified a line of research that focuses on the notion of empirical risk parity~\cite{tran2021differentially,noe2022exploring,xu2019achieving,esipova2022disparate}. However, the relationship between risk parity and other fairness notions remains unclear. The predictive performance of a model is not driven solely by the loss function, as instances with high loss can still be correctly predicted. This puts into question the presumed link between risk parity and other fairness notions during the deployment of the models.

The definition of fairness varies based on how datasets are grouped. While many studies use predefined binary groupings, real-world datasets often involve intersections of multiple protected attributes, such as gender, race, and religion~\cite{compas_dataset,adult_dataset,pew_religious_landscape,gender_shades}. These intersections create complex, multi-dimensional groupings, which can influence factors like data imbalance and group proximity to the decision boundary. This raises key questions: Do complex groupings reveal distinct disparities? Can addressing fairness for one grouping worsen disparities for another? These aspects remain underexplored.

\subsection{A Caveat on Alternatives to DP-SGD for Fairness}

In Section \ref{sec:factors}, we review literature on approaches that aim to balance privacy and fairness, focusing on DP techniques beyond DP-SGD. While these alternatives show promise, their ability to provide privacy guarantees equivalent to DP-SGD and resist inference attacks is uncertain. For example, Noe et al.\ demonstrate that while applying DP to PCA has minimal impact on fairness outcomes, their experiments exposed vulnerabilities that allow attackers to infer information about the training dataset in models reduced through DP-PCA, raising privacy concerns~\cite{noe2022exploring}. This suggests that while some alternative DP algorithms may avoid the disparate impacts seen with DP-SGD, they might undermine privacy in downstream tasks. Therefore, we advise caution when adopting DP algorithms that claim to improve fairness, recommending a thorough evaluation of robustness and a comparison to DP-SGD.

\section{Conclusion}
We have comprehensively reviewed the literature to identify all the contributing factors reported to exacerbate performance disparities in DP-based PPML. We propose a taxonomy to categorize these factors based on their location within the ML pipeline that sheds light on the importance of each factor. Furthermore, we have conducted a causal analysis of the necessity and sufficiency of these factors in causing the observed disparate effects of DP. Our findings suggest that a combination of a small training dataset size and disparate group distances to the decision boundary may be sufficient for the exacerbation of disparities to arise under the implementation of DP. We have also reviewed existing mitigation strategies, highlighting their limitations. Finally, we identify several key research gaps and propose promising potential directions for future research in this area.

\section*{Acknowledgments}
We would like to thank the anonymous reviewers for their constructive comments and suggestions to the paper during the peer review process. We also extend our appreciation to Rik Sarkar and Bogdan Kulynych for their valuable feedback on the project. Marc Juarez is supported by a Google Research Scholar Award.

\bibliographystyle{IEEEtran}
\bibliography{reference.bib}

\newpage
\appendices
\section{Google Scholar Search Query}
\label{appendix_googlescholar}

In this section, we provide the details about the construction of the Google Scholar search query used to find candidate studies for this survey. 

Table~\ref{tab:googlescholarquery} provides an itemized description of the search query, which can also be formulated in one line as:

\begin{center}
\texttt{\small "differential privacy" AND 
"fairness" AND 
("machine learning" OR "deep learning") AND 
(intitle:"private" OR intitle:"privacy" OR intitle:"DP" OR intitle:"LDP" OR intitle:"PATE" OR intitle:"Randomized Response") AND 
(intitle:"fair" OR intitle:"unfair" OR intitle:"fairness" OR intitle:"unfairness" OR intitle:"disparate" OR intitle:"disparity") AND 
-intitle:"federated" -intitle:"generative" -intitle:"GAN"
}    
\end{center}


The rationale for the breakdown in Table~\ref{tab:googlescholarquery} is as follows: We aim to locate papers that specifically address fairness issues in ML systems that implement DP. To achieve this, the content must include the terms \texttt{differential privacy}, \texttt{fairness}, and \texttt{machine learning}. However, a query with only these terms is too broad, returning thousands of results, most of them out of the scope of our survey. To refine the query, we observed that papers within this research area typically also have these concepts explicitly included in their titles. Papers that conduct such research usually have titles that contain DP or privacy-related terms such as \texttt{differential privacy} or \texttt{privacy}, alongside fairness-related terms like \texttt{fairness} or \texttt{disparity}. Note that although we are specifically interested in ML-related works, not all such papers include ML-related terms in their titles, so we did not impose this as a strict requirement. This constraint efficiently narrows down the search result.

Additionally, we want to exclude works outside our scope, such as those focused on federated learning or generative models, which is why we included several exclusion terms in the search.

\begin{table*}[ht]
\caption{Google Scholar search query breakdown}
\begin{center}
\begin{tabular}{l|l}
\toprule
\textbf{Category}                                                   & \textbf{Query Terms}                                                                                             \\ \midrule
\multirow[t]{2}{*}{Privacy (in content)}                              & {\footnotesize\texttt{"differential privacy" AND}}                \\ \midrule
\multirow[t]{2}{*}{Privacy (in title)} & \parbox[t]{15cm}{\footnotesize\texttt{(intitle:"private" OR intitle:"privacy" OR intitle:"DP" OR intitle:"LDP" OR intitle:"PATE" OR intitle:"Randomized Response") AND}} \\ \midrule
\multirow[t]{2}{*}{Fairness (in content)}                             & {\footnotesize\texttt{"fairness" AND}}                                \\ \midrule
\multirow[t]{2}{*}{Fairness (in title)} & \parbox[t]{15cm}{\footnotesize\texttt{(intitle:"fair" OR intitle:"unfair" OR intitle:"fairness" OR intitle:"unfairness" OR intitle:"disparate" OR intitle:"disparity") AND}} \\ \midrule
\multirow[t]{2}{*}{ML (in content)}                             & {\footnotesize\texttt{("machine learning" OR "deep learning") AND}}                                                            \\ \midrule
\multirow[t]{2}{*}{Exclusion (in title)}                                     & {\footnotesize\texttt{-intitle:"federated" -intitle:"generative" -intitle:"GAN"}}                                                                                \\ \bottomrule
\end{tabular}
\end{center}
\label{tab:googlescholarquery}
\end{table*}

\section{Evaluation \& Experiment Details of Each Study}
\label{details}

We summarize the evaluation and experimental details of each study in Table~\ref{details_table}, sorted by order of appearance in the text. This table provides key information for interested readers, including the fairness definitions, the DP techniques, the ML algorithms, and the data modalities considered in each study.

Next, we detail some of the aspects of the table:

\begin{itemize}
    \item The entry labeled ``Theory'' under the ``DP Technique'' category refers to a study that is purely theoretical and does not involve empirical experiments. This study by Cummings et al.\ focuses on a theoretical examination of DP and its implications for fairness \cite{cummings2019compatibility}. As it does not employ any experimental DP techniques, it is listed separately from the DP techniques included in our survey.
    
    \item The ``Transformers'' entry under the ``ML Algorithm'' category refers to the use of Transformer models \cite{lin2022survey}. These models were originally developed for NLP tasks, though they are increasingly applied in computer vision as well. Specifically, Noe et al.~\cite{noe2022exploring} used BERT~\cite{devlin2018bert} models, while Srivastava et al.~\cite{srivastava2024amplifying} employed DistilGPT2~\cite{sanh2019distilbert}, a model from the GPT family.
    
    \item Upon reviewing the table, we noticed a trend: studies conducted before 2023 typically analyzed only one, or at most two, fairness notions at a time. However, since 2023, there has been a significant increase in studies that analyze multiple fairness notions (three or more) simultaneously. This shift aligns with our suggestion in section \ref{sec:discussion} to include multiple fairness notions in the evaluation methodology; this would enable a more comprehensive understanding and aid comparability across works.
\end{itemize}

We also provide an explanation of the acronyms used in Table~\ref{details_table} below:

\begin{tcolorbox}[title=Glossary, rounded corners]
\textbf{Fairness Notion (also defined in Section \ref{fairness_notions})}

\begin{acronym}[longest]
    \setlength{\itemsep}{0pt}
    \acro{AP}{Accuracy Parity}
    \acro{DEP}{Demographic Parity}
    \acro{EOP}{Equal Opportunity}
    \acro{EOD}{Equalized Odds}
    \acro{PEP}{Predictive Equality Parity}
    \acro{PRP}{Predictive Rate Parity}
    \acro{RP}{Risk Parity}
\end{acronym}

\textbf{DP Framework (also defined in Section \ref{privacy_in_ml})}

\begin{acronym}[longest]
    \setlength{\itemsep}{0pt}
    \acro{GDP}{Global DP}
    \acro{LDP}{Local DP}
\end{acronym}

\vspace{-5pt}
\textbf{GDP: others} includes: PATE, DP-FedAvg, Functional Mechanism, and Output Perturbation. \textbf{LDP} includes: DP-PCA, and Randomized Response.
\vspace{10pt}

\textbf{ML Algorithm}

\begin{acronym}[longest]
    \setlength{\itemsep}{0pt}
    \acro{NN}{Neural Network}
    \acro{LR}{Logistic Regression}
    \acro{Tree}{Decision Tree-based models}
    \acro{NB}{Naive Bayes}
\end{acronym}

\vspace{-5pt}
\textbf{NN} includes: shallow networks like Multi-Layer Perceptron (MLP), deep networks like ResNet50, and recurrent networks like Long Short-Term Memory (LSTM). \textbf{Tree} includes: Random Forests, and Light Gradient Boosting Machine (LightGBM).

\end{tcolorbox}

In our study, we recognize the diversity present in the tasks, datasets, ML models, and DP techniques employed across the body of work we analyzed. This diversity reflects the evolving nature of research in DP and fairness, a field still in its early stages. Across the 20 studies included in our scope, there is considerable variation in experimental settings, fairness notions, and DP mechanisms. We have documented this diversity comprehensively in Table~\ref{details_table} and provided detailed descriptions of the tasks and models used in each study to offer readers a nuanced understanding of the current landscape.

Despite the variation, certain overlaps in experimental conditions allow for meaningful comparisons, even when studies approach similar setups from different perspectives. For instance, studies \cite{xu2020removing} and \cite{esipova2022disparate} both examine logistic regression with DP-SGD on the Dutch dataset to analyze accuracy disparity. However, their focuses diverge: \cite{xu2020removing} develops mitigation methods, specifically a group-wise adaptive clipping mechanism for DP-SGD, while \cite{esipova2022disparate} highlights the role of gradient clipping and its impact on fairness, particularly regarding the directional error induced by the clipping. These differences complement rather than conflict with one another. Similarly, studies \cite{sanyal2022unfair}, \cite{farrand2020neither}, and \cite{esipova2022disparate} investigate accuracy disparity in neural networks trained on the CelebA dataset with DP-SGD, but their analyses vary—\cite{sanyal2022unfair} and \cite{farrand2020neither} explore the effects of data distributional properties and group imbalance, whereas \cite{esipova2022disparate} examines gradient clipping as a contributing factor. On the MNIST dataset, studies \cite{uniyal2021dp}, \cite{xu2020removing}, and \cite{esipova2022disparate} offer further complementary insights by addressing distinct goals, such as comparing DP-SGD with PATE (\cite{uniyal2021dp}), developing mitigation techniques \cite{xu2020removing}, and analyzing gradient clipping effects \cite{esipova2022disparate}.

This diversity, while posing challenges to direct comparability, enriches the field by providing multifaceted insights. The variation across studies enables high-level causal analysis, allowing us to identify contributing factors and potentially \textit{necessary} factors to DP-induced unfairness. Counterexamples help demonstrate the insufficiency of certain factors without requiring strict comparability, while variation across settings supports the inference of necessity. Where evidence is less definitive, we qualify the strength of our claims, as detailed in Table~\ref{tab:listfactors}.

By presenting this diversity alongside shared insights, our work highlights how complementary perspectives converge to form a coherent understanding of DP-induced unfairness.

\begin{landscape}
\begin{table}[]
\caption{Experimental details of each study analyzed in our survey, in the order of citation appearance in the main text. The details include the fairness notions, the DP techniques, the ML algorithms, and the data modalities considered in each study.}
\label{details_table}

\begin{tabular}{@{}|l|l|ccccccc|cccc|ccccc|ccc|@{}}
\toprule
\multicolumn{1}{|c|}{\multirow{2}{*}{\textbf{Work}}} & \multicolumn{1}{c|}{\multirow{2}{*}{\textbf{Year}}} & \multicolumn{7}{c|}{\textbf{Fairness Notion}}                                                                                                                                                                                      & \multicolumn{4}{c|}{\textbf{DP Technique}}                                                                                                  & \multicolumn{5}{c|}{\textbf{ML Algorithm}}                                                                                                                          & \multicolumn{3}{c|}{\textbf{Data Modality}}                                                      \\ \cmidrule(l){3-21} 
\multicolumn{1}{|c|}{}                               & \multicolumn{1}{c|}{}                               & \multicolumn{1}{c|}{\textbf{AP}} & \multicolumn{1}{c|}{\textbf{DEP}} & \multicolumn{1}{c|}{\textbf{EOP}} & \multicolumn{1}{c|}{\textbf{EOD}} & \multicolumn{1}{c|}{\textbf{PEP}} & \multicolumn{1}{c|}{\textbf{PRP}} & \textbf{RP} & \multicolumn{1}{c|}{\textbf{GDP: DP-SGD}} & \multicolumn{1}{c|}{\textbf{GDP: others}} & \multicolumn{1}{c|}{\textbf{LDP}} & \textbf{Theory} & \multicolumn{1}{c|}{\textbf{NN}} & \multicolumn{1}{c|}{\textbf{LR}} & \multicolumn{1}{c|}{\textbf{Tree}} & \multicolumn{1}{c|}{\textbf{NB}} & \textbf{Transformers} & \multicolumn{1}{c|}{\textbf{Tabular}} & \multicolumn{1}{c|}{\textbf{Image}} & \textbf{Text} \\ \midrule
\cite{bagdasaryan2019differential}                   & 2019                                                & \multicolumn{1}{c|}{$\newmoon$}   & \multicolumn{1}{c|}{}             & \multicolumn{1}{c|}{}             & \multicolumn{1}{c|}{}             & \multicolumn{1}{c|}{}             & \multicolumn{1}{c|}{}             &             & \multicolumn{1}{c|}{$\newmoon$}            & \multicolumn{1}{c|}{}                     & \multicolumn{1}{c|}{}             &                 & \multicolumn{1}{c|}{$\newmoon$}   & \multicolumn{1}{c|}{}            & \multicolumn{1}{c|}{}              & \multicolumn{1}{c|}{}            &                       & \multicolumn{1}{c|}{}                & \multicolumn{1}{c|}{$\newmoon$}       & $\newmoon$    \\ \midrule
\cite{xu2020removing}                                & 2020                                                & \multicolumn{1}{c|}{$\newmoon$}   & \multicolumn{1}{c|}{}             & \multicolumn{1}{c|}{}             & \multicolumn{1}{c|}{}             & \multicolumn{1}{c|}{}             & \multicolumn{1}{c|}{}             &             & \multicolumn{1}{c|}{$\newmoon$}            & \multicolumn{1}{c|}{}                     & \multicolumn{1}{c|}{}             &                 & \multicolumn{1}{c|}{$\newmoon$}   & \multicolumn{1}{c|}{$\newmoon$}   & \multicolumn{1}{c|}{}              & \multicolumn{1}{c|}{}            &                       & \multicolumn{1}{c|}{$\newmoon$}       & \multicolumn{1}{c|}{$\newmoon$}       &              \\ \midrule
\cite{tran2021differentially}                        & 2021                                                & \multicolumn{1}{c|}{}            & \multicolumn{1}{c|}{}             & \multicolumn{1}{c|}{}             & \multicolumn{1}{c|}{}             & \multicolumn{1}{c|}{}             & \multicolumn{1}{c|}{}             & $\newmoon$   & \multicolumn{1}{c|}{$\newmoon$}            & \multicolumn{1}{c|}{$\newmoon$}            & \multicolumn{1}{c|}{}             &                 & \multicolumn{1}{c|}{$\newmoon$}   & \multicolumn{1}{c|}{}            & \multicolumn{1}{c|}{}              & \multicolumn{1}{c|}{}            &                       & \multicolumn{1}{c|}{$\newmoon$}       & \multicolumn{1}{c|}{}                &              \\ \midrule
\cite{esipova2022disparate}                          & 2022                                                & \multicolumn{1}{c|}{$\newmoon$}   & \multicolumn{1}{c|}{}             & \multicolumn{1}{c|}{}             & \multicolumn{1}{c|}{}             & \multicolumn{1}{c|}{}             & \multicolumn{1}{c|}{}             & $\newmoon$   & \multicolumn{1}{c|}{$\newmoon$}            & \multicolumn{1}{c|}{}                     & \multicolumn{1}{c|}{}             &                 & \multicolumn{1}{c|}{$\newmoon$}   & \multicolumn{1}{c|}{$\newmoon$}   & \multicolumn{1}{c|}{}              & \multicolumn{1}{c|}{}            &                       & \multicolumn{1}{c|}{$\newmoon$}       & \multicolumn{1}{c|}{$\newmoon$}       &              \\ \midrule
\cite{srivastava2024amplifying}                      & 2024                                                & \multicolumn{1}{c|}{}            & \multicolumn{1}{c|}{}             & \multicolumn{1}{c|}{}             & \multicolumn{1}{c|}{}             & \multicolumn{1}{c|}{}             & \multicolumn{1}{c|}{}             &             & \multicolumn{1}{c|}{$\newmoon$}            & \multicolumn{1}{c|}{}                     & \multicolumn{1}{c|}{}             &                 & \multicolumn{1}{c|}{}            & \multicolumn{1}{c|}{}            & \multicolumn{1}{c|}{}              & \multicolumn{1}{c|}{}            & $\newmoon$             & \multicolumn{1}{c|}{}                & \multicolumn{1}{c|}{}                & $\newmoon$    \\ \midrule
\cite{liu2022mitigating}                             & 2022                                                & \multicolumn{1}{c|}{$\newmoon$}   & \multicolumn{1}{c|}{}             & \multicolumn{1}{c|}{}             & \multicolumn{1}{c|}{}             & \multicolumn{1}{c|}{}             & \multicolumn{1}{c|}{}             &             & \multicolumn{1}{c|}{$\newmoon$}            & \multicolumn{1}{c|}{$\newmoon$}            & \multicolumn{1}{c|}{}             &                 & \multicolumn{1}{c|}{$\newmoon$}   & \multicolumn{1}{c|}{$\newmoon$}   & \multicolumn{1}{c|}{}              & \multicolumn{1}{c|}{}            &                       & \multicolumn{1}{c|}{$\newmoon$}       & \multicolumn{1}{c|}{$\newmoon$}       &              \\ \midrule
\cite{uniyal2021dp}                                  & 2021                                                & \multicolumn{1}{c|}{$\newmoon$}   & \multicolumn{1}{c|}{}             & \multicolumn{1}{c|}{}             & \multicolumn{1}{c|}{}             & \multicolumn{1}{c|}{}             & \multicolumn{1}{c|}{}             &             & \multicolumn{1}{c|}{$\newmoon$}            & \multicolumn{1}{c|}{$\newmoon$}            & \multicolumn{1}{c|}{}             &                 & \multicolumn{1}{c|}{$\newmoon$}   & \multicolumn{1}{c|}{}            & \multicolumn{1}{c|}{}              & \multicolumn{1}{c|}{}            &                       & \multicolumn{1}{c|}{}                & \multicolumn{1}{c|}{$\newmoon$}       &              \\ \midrule
\cite{xu2019achieving}                               & 2019                                                & \multicolumn{1}{c|}{}            & \multicolumn{1}{c|}{}             & \multicolumn{1}{c|}{}             & \multicolumn{1}{c|}{}             & \multicolumn{1}{c|}{}             & \multicolumn{1}{c|}{}             & $\newmoon$   & \multicolumn{1}{c|}{}                     & \multicolumn{1}{c|}{$\newmoon$}            & \multicolumn{1}{c|}{}             &                 & \multicolumn{1}{c|}{}            & \multicolumn{1}{c|}{$\newmoon$}   & \multicolumn{1}{c|}{}              & \multicolumn{1}{c|}{}            &                       & \multicolumn{1}{c|}{$\newmoon$}       & \multicolumn{1}{c|}{}                &              \\ \midrule
\cite{carey2023randomized}                           & 2023                                                & \multicolumn{1}{c|}{$\newmoon$}   & \multicolumn{1}{c|}{}             & \multicolumn{1}{c|}{}             & \multicolumn{1}{c|}{}             & \multicolumn{1}{c|}{}             & \multicolumn{1}{c|}{}             &             & \multicolumn{1}{c|}{}                     & \multicolumn{1}{c|}{}                     & \multicolumn{1}{c|}{$\newmoon$}    &                 & \multicolumn{1}{c|}{$\newmoon$}   & \multicolumn{1}{c|}{$\newmoon$}   & \multicolumn{1}{c|}{$\newmoon$}     & \multicolumn{1}{c|}{$\newmoon$}   &                       & \multicolumn{1}{c|}{$\newmoon$}       & \multicolumn{1}{c|}{$\newmoon$}       &              \\ \midrule
\cite{mangold2023differential}                       & 2023                                                & \multicolumn{1}{c|}{$\newmoon$}   & \multicolumn{1}{c|}{$\newmoon$}    & \multicolumn{1}{c|}{$\newmoon$}    & \multicolumn{1}{c|}{$\newmoon$}    & \multicolumn{1}{c|}{}             & \multicolumn{1}{c|}{}             &             & \multicolumn{1}{c|}{$\newmoon$}            & \multicolumn{1}{c|}{$\newmoon$}            & \multicolumn{1}{c|}{}             &                 & \multicolumn{1}{c|}{}            & \multicolumn{1}{c|}{$\newmoon$}   & \multicolumn{1}{c|}{}              & \multicolumn{1}{c|}{}            &                       & \multicolumn{1}{c|}{$\newmoon$}       & \multicolumn{1}{c|}{$\newmoon$}       &              \\ \midrule
\cite{makhlouf2024impact}                            & 2024                                                & \multicolumn{1}{c|}{$\newmoon$}   & \multicolumn{1}{c|}{$\newmoon$}    & \multicolumn{1}{c|}{$\newmoon$}    & \multicolumn{1}{c|}{}             & \multicolumn{1}{c|}{$\newmoon$}    & \multicolumn{1}{c|}{$\newmoon$}    &             & \multicolumn{1}{c|}{}                     & \multicolumn{1}{c|}{}                     & \multicolumn{1}{c|}{$\newmoon$}    &                 & \multicolumn{1}{c|}{}            & \multicolumn{1}{c|}{}            & \multicolumn{1}{c|}{$\newmoon$}     & \multicolumn{1}{c|}{}            &                       & \multicolumn{1}{c|}{$\newmoon$}       & \multicolumn{1}{c|}{}                &              \\ \midrule
\cite{noe2022exploring}                              & 2022                                                & \multicolumn{1}{c|}{}            & \multicolumn{1}{c|}{}             & \multicolumn{1}{c|}{}             & \multicolumn{1}{c|}{$\newmoon$}    & \multicolumn{1}{c|}{}             & \multicolumn{1}{c|}{}             & $\newmoon$   & \multicolumn{1}{c|}{$\newmoon$}            & \multicolumn{1}{c|}{}                     & \multicolumn{1}{c|}{$\newmoon$}    &                 & \multicolumn{1}{c|}{$\newmoon$}   & \multicolumn{1}{c|}{$\newmoon$}   & \multicolumn{1}{c|}{}              & \multicolumn{1}{c|}{}            & $\newmoon$             & \multicolumn{1}{c|}{}                & \multicolumn{1}{c|}{}                & $\newmoon$    \\ \midrule
\cite{makhlouf2024systematic}                        & 2024                                                & \multicolumn{1}{c|}{}            & \multicolumn{1}{c|}{$\newmoon$}    & \multicolumn{1}{c|}{$\newmoon$}    & \multicolumn{1}{c|}{}             & \multicolumn{1}{c|}{}             & \multicolumn{1}{c|}{}             &             & \multicolumn{1}{c|}{}                     & \multicolumn{1}{c|}{}                     & \multicolumn{1}{c|}{$\newmoon$}    &                 & \multicolumn{1}{c|}{}            & \multicolumn{1}{c|}{}            & \multicolumn{1}{c|}{$\newmoon$}     & \multicolumn{1}{c|}{}            &                       & \multicolumn{1}{c|}{$\newmoon$}       & \multicolumn{1}{c|}{}                &              \\ \midrule
\cite{de2023empirical}                               & 2023                                                & \multicolumn{1}{c|}{}            & \multicolumn{1}{c|}{$\newmoon$}    & \multicolumn{1}{c|}{}             & \multicolumn{1}{c|}{$\newmoon$}    & \multicolumn{1}{c|}{}             & \multicolumn{1}{c|}{$\newmoon$}    &             & \multicolumn{1}{c|}{$\newmoon$}            & \multicolumn{1}{c|}{}                     & \multicolumn{1}{c|}{}             &                 & \multicolumn{1}{c|}{$\newmoon$}   & \multicolumn{1}{c|}{}            & \multicolumn{1}{c|}{}              & \multicolumn{1}{c|}{}            &                       & \multicolumn{1}{c|}{$\newmoon$}       & \multicolumn{1}{c|}{}                &              \\ \midrule
\cite{cummings2019compatibility}                     & 2019                                                & \multicolumn{1}{c|}{}            & \multicolumn{1}{c|}{}             & \multicolumn{1}{c|}{$\newmoon$}    & \multicolumn{1}{c|}{}             & \multicolumn{1}{c|}{}             & \multicolumn{1}{c|}{}             &             & \multicolumn{1}{c|}{}                     & \multicolumn{1}{c|}{}                     & \multicolumn{1}{c|}{}             & $\newmoon$       & \multicolumn{1}{c|}{}            & \multicolumn{1}{c|}{}            & \multicolumn{1}{c|}{}              & \multicolumn{1}{c|}{}            &                       & \multicolumn{1}{c|}{}                & \multicolumn{1}{c|}{}                &              \\ \midrule
\cite{arasteh2023private}                            & 2023                                                & \multicolumn{1}{c|}{$\newmoon$}   & \multicolumn{1}{c|}{}             & \multicolumn{1}{c|}{}             & \multicolumn{1}{c|}{}             & \multicolumn{1}{c|}{}             & \multicolumn{1}{c|}{}             &             & \multicolumn{1}{c|}{$\newmoon$}            & \multicolumn{1}{c|}{}                     & \multicolumn{1}{c|}{}             &                 & \multicolumn{1}{c|}{$\newmoon$}   & \multicolumn{1}{c|}{}            & \multicolumn{1}{c|}{}              & \multicolumn{1}{c|}{}            &                       & \multicolumn{1}{c|}{}                & \multicolumn{1}{c|}{$\newmoon$}       &              \\ \midrule
\cite{sanyal2022unfair}                              & 2022                                                & \multicolumn{1}{c|}{$\newmoon$}   & \multicolumn{1}{c|}{}             & \multicolumn{1}{c|}{}             & \multicolumn{1}{c|}{}             & \multicolumn{1}{c|}{}             & \multicolumn{1}{c|}{}             &             & \multicolumn{1}{c|}{$\newmoon$}            & \multicolumn{1}{c|}{}                     & \multicolumn{1}{c|}{}             &                 & \multicolumn{1}{c|}{$\newmoon$}   & \multicolumn{1}{c|}{}            & \multicolumn{1}{c|}{$\newmoon$}     & \multicolumn{1}{c|}{}            &                       & \multicolumn{1}{c|}{$\newmoon$}       & \multicolumn{1}{c|}{$\newmoon$}       &              \\ \midrule
\cite{farrand2020neither}                            & 2020                                                & \multicolumn{1}{c|}{$\newmoon$}   & \multicolumn{1}{c|}{}             & \multicolumn{1}{c|}{}             & \multicolumn{1}{c|}{}             & \multicolumn{1}{c|}{}             & \multicolumn{1}{c|}{}             &             & \multicolumn{1}{c|}{$\newmoon$}            & \multicolumn{1}{c|}{}                     & \multicolumn{1}{c|}{}             &                 & \multicolumn{1}{c|}{$\newmoon$}   & \multicolumn{1}{c|}{}            & \multicolumn{1}{c|}{}              & \multicolumn{1}{c|}{}            &                       & \multicolumn{1}{c|}{}                & \multicolumn{1}{c|}{$\newmoon$}       &              \\ \midrule
\cite{tran2023fairdp}                                & 2023                                                & \multicolumn{1}{c|}{}            & \multicolumn{1}{c|}{$\newmoon$}    & \multicolumn{1}{c|}{$\newmoon$}    & \multicolumn{1}{c|}{$\newmoon$}    & \multicolumn{1}{c|}{}             & \multicolumn{1}{c|}{}             &             & \multicolumn{1}{c|}{$\newmoon$}            & \multicolumn{1}{c|}{}                     & \multicolumn{1}{c|}{}             &                 & \multicolumn{1}{c|}{$\newmoon$}   & \multicolumn{1}{c|}{}            & \multicolumn{1}{c|}{}              & \multicolumn{1}{c|}{}            &                       & \multicolumn{1}{c|}{$\newmoon$}       & \multicolumn{1}{c|}{}                &              \\ \bottomrule
\end{tabular}
\end{table}
\end{landscape}

\end{document}